%% file: main.tex
\documentclass[10pt,twocolumn,letterpaper]{article}
\usepackage[pagenumbers]{iccv} 
\newif\ifemail
\newif\ifchecklist
\newif\ifappendix
\newif\ifbanner
\usepackage{soul}
\usepackage{graphicx}
\usepackage{amsmath}
\usepackage{bbm}
\usepackage{amsthm}
\usepackage{booktabs}
\usepackage{algorithm}
\usepackage{tabularx}
\usepackage{tabu}
\usepackage{array}
\usepackage{times}
\usepackage{amssymb}
\usepackage{multirow}
\usepackage{wrapfig}
\usepackage{makecell}
\usepackage{colortbl}
\usepackage{bbding} 
\usepackage{color}
\usepackage{makecell}
\usepackage{pifont} 
\newcommand{\cmark}{\ding{51}}%

\usepackage{setspace}
\usepackage{algorithmicx}

\definecolor{Blue}{RGB}{0, 0, 255}
\definecolor{Aquamarine}{RGB}{127, 255, 212}
\definecolor{Sepia}{RGB}{112, 66, 20}
\definecolor{BrickRed}{RGB}{203, 65, 84}
\colorlet{my-red}{BrickRed!90!Sepia}
\colorlet{my-blue}{Aquamarine!30!Blue}
\input{newcommand}

\usepackage{tikz}
\usepackage{mathtools}
\usepackage[normalem]{ulem}  %
\usepackage{pgfplots,pgfplotstable}
\usepgfplotslibrary{fillbetween}
\usepackage{calc}
\usepackage{pbox}
\usepackage[noend]{algpseudocode}
\usepackage{afterpage}


\definecolor{cvprblue}{rgb}{0.21,0.49,0.74}
\definecolor{aliceblue}{RGB}{240, 230, 140}
\newcommand{\CC}{\cellcolor{aliceblue}}
\definecolor{babyblue}{RGB}{255, 255, 255}

\usepackage[pagebackref,breaklinks,colorlinks,allcolors=cvprblue]{hyperref}

\def\paperID{8400} 
\def\confName{ICCV}
\def\confYear{2025}

\title{Bag of Design Choices for Inference of High-Resolution\\Masked Generative Transformer}

\author{Shitong Shao$^1$ \qquad Zikai Zhou$^{1,\diamondsuit}$ \qquad Tian Ye$^{1}$ \qquad Lichen Bai$^{1}$ \qquad Zhiqiang Xu$^{2}$\qquad Zeke Xie$^{1,*}$ \\
  $^1$Hong Kong University of Science and Technology (Guangzhou) \\  
  $^2$Mohamed bin Zayed University of Artificial Intelligence \\
{\tt\small \{sshao213,zikaizhou,lichenbai,zekexie\}@hkust-gz.edu.cn} \\
{\tt\small tye610@connnect.hkust-gz.edu.cn \qquad zhiqiang.xu@mbzuai.ac.ae} \\ 
{\tt \small $^\diamondsuit$:Equal Contribution \qquad $*$:Corresponding author}}

\begin{document}
\maketitle
\input{sec/abstract}
\input{sec/intro}
\input{sec/related_work}
\input{sec/method}
\input{sec/experiment}
\input{sec/conclusion}


{
    \small
    \bibliographystyle{ieeenat_fullname}
    \bibliography{main}
}

\input{sec/suppl}

\end{document}

%% file: newcommand.tex
\def\clap#1{\hbox to 0pt{\hss #1\hss}}%

\newcommand{\titlerowww}[1]{\makebox[0mm][l]{{\bf #1}}\\}

\newcommand{\ve}[1]{\mathbf{#1}}

\definecolor{C0}{rgb}{0.121569, 0.466667, 0.705882}
\definecolor{C1}{rgb}{1.000000, 0.498039, 0.054902}
\definecolor{C2}{rgb}{0.172549, 0.627451, 0.172549}
\definecolor{C3}{rgb}{0.839216, 0.152941, 0.156863}
\definecolor{C4}{rgb}{0.580392, 0.403922, 0.741176}
\definecolor{C5}{rgb}{0.549020, 0.337255, 0.294118}
\definecolor{C6}{rgb}{0.890196, 0.466667, 0.760784}
\definecolor{C7}{rgb}{0.498039, 0.498039, 0.498039}
\definecolor{C8}{rgb}{0.737255, 0.741176, 0.133333}
\definecolor{C9}{rgb}{0.090196, 0.745098, 0.811765}

%% file: sec/abstract.tex
\begin{abstract}
Text-to-image diffusion models (DMs) develop at an unprecedented pace, supported by thorough theoretical exploration and empirical analysis. Unfortunately, the discrepancy between DMs and autoregressive models (ARMs) complicates the path toward achieving the goal of unified vision and language generation. Recently, the masked generative Transformer (MGT) serves as a promising intermediary between DM and ARM by predicting randomly masked image tokens (\textit{i.e.}, masked image modeling), combining the efficiency of DM with the discrete token nature of ARM. However, we find that the comprehensive analyses regarding the inference for MGT are virtually non-existent, and thus we aim to present positive design choices to fill this gap. We propose and redesign a set of enhanced inference techniques tailored for MGT, providing a detailed analysis of their performance. Additionally, we explore several DM-based approaches aimed at accelerating the sampling process on MGT. Extensive experiments and empirical analyses on the recent SOTA MGT, such as MaskGIT and Meissonic lead to concrete and effective design choices, and these design choices can be merged to achieve further performance gains. For instance, in terms of enhanced inference, we achieve winning rates of approximately 70\% compared to vanilla sampling on HPS v2 with Meissonic-1024$\times$1024.
\end{abstract}

%% file: sec/intro.tex
\section{Introduction}
\label{sec:intro}

\begin{figure*}[t]
\vspace{-8pt}
\centering
\includegraphics[width=0.95\textwidth,trim={0cm 0cm 0cm 0cm},clip]{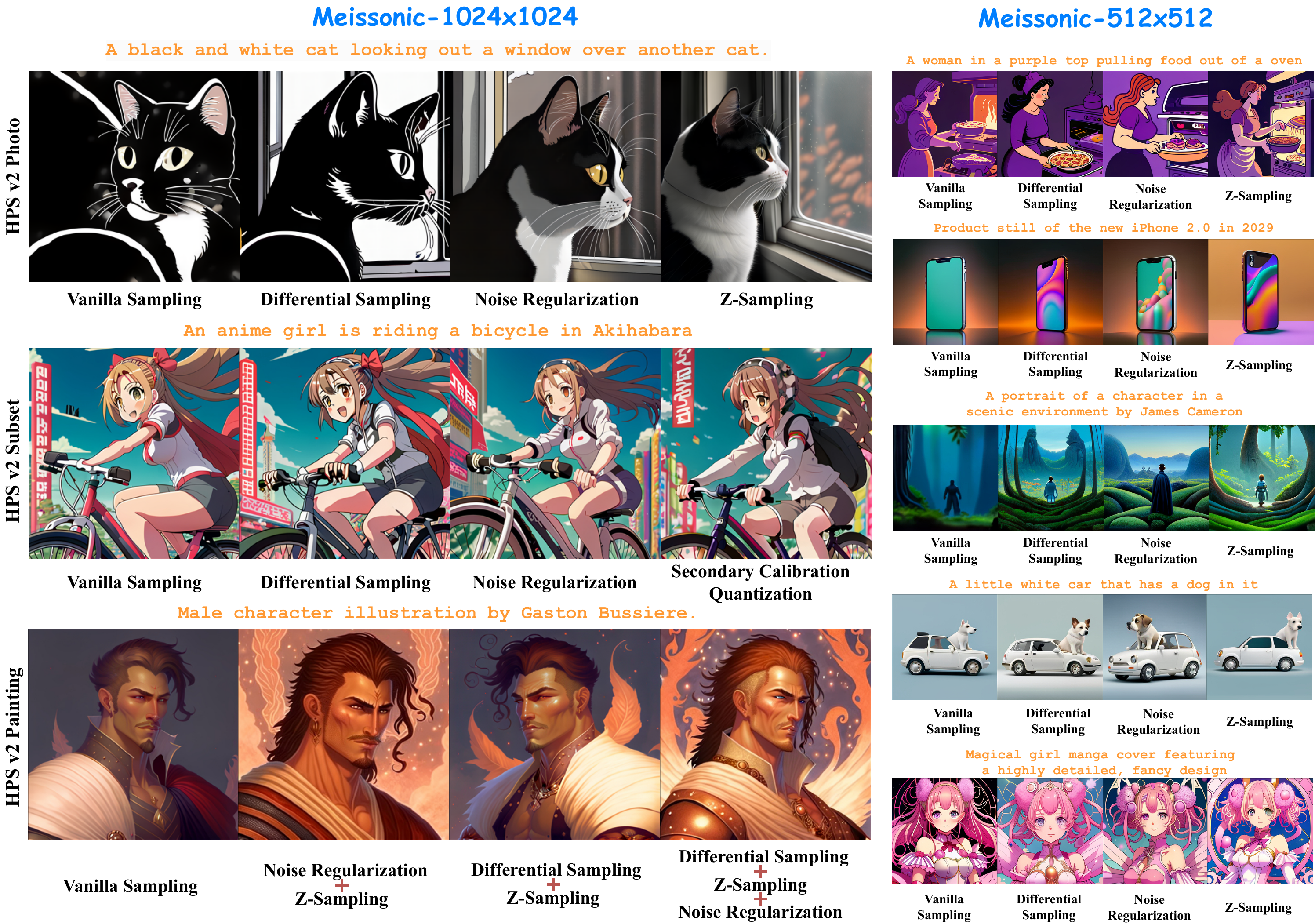}
\vspace{-6pt}
\caption{Visualization of our design choices on Meissonic-512$\times$512 and Meissonic-1024$\times$1024. Noise regularization, differential sampling, and Z-Sampling significantly improve both visual quality and semantic faithfulness. Additionally, our quantization method secondary calibration quantization (SCQ) can reduce the memory footprint without significant performance degradation.}
\vspace{-12pt}
\label{figure:visualization}
\end{figure*}
The rapid development of generative models has successfully sparked a deep learning innovations in both computer vision and natural language processing. The emergence of large language models (LLMs) in natural language processing, along with their strong generalization across domains and tasks~\citep{llama,gemma,strawberry}, benefits from the autoregressive model (ARM) with the Transformer decoder block~\citep{Transformer}. By contrast, the dominant paradigm in text-to-image (T2I) synthesis is the diffusion model (DM), which employs a multi-step denoising process to synthesize high-quality images from Gaussian noise~\citep{ddpm_begin,sde,ddim,liu2024alignment,shao2023catch,shao2023diffuseexpand}. The significant variability in training and inference between ARMs and DMs hinders the unification of generative paradigms in computer vision and natural language processing. The recent accomplishments of some ARMs in visual generation, such as LlamaGen~\citep{llamagen}, Lumina-mGPT~\citep{luminia_mgpt}, and Fluid~\citep{fluid}, indicates that DMs are not the sole option for achieving success in image generation. This paradigm synthesizes extremely high-quality images but requires hundreds or even thousands of function evaluations (NFEs) to synthesize a single image~\citep{jacobidecoding}. Instead, masked generative Transformers (MGTs)~\citep{maskgit} predict multiple masked tokens in each forward pass, resulting a trade-off between DMs and ARMs. This approach preserves the efficiency of DMs while stabilizing the transformation of images into discrete tokens, thus aligning with the part characterization of LLM~\citep{sennrich2015neural}.

A recent MGT, Meissonic~\citep{meissonic}, achieved high-quality image synthesis at the 1024$\times$1024 resolution for the first time, setting a new state-of-the-art performance on HPS v2~\citep{wu2023human} and outperforming SD XL~\citep{SDXL} by a margin of 0.69. This phenomenon substantiates MGT's capability to synthesize high-resolution images and suggests the potential for developing a commercial-grade generative model, such as FLUX~\citep{FLUX}. Unfortunately, compared to the extensive theoretical research and empirical analysis in the DM field, scholarly exploration and understanding of high-resolution MGTs remain unexplored, hindering the further development of MGT in both training and inference~\citep{maskgit,muse,meissonic}.

\begin{figure*}[!t]
\vspace{-8pt}
\centering
\includegraphics[width=0.89\textwidth,trim={0cm 0cm 0cm 0cm},clip]{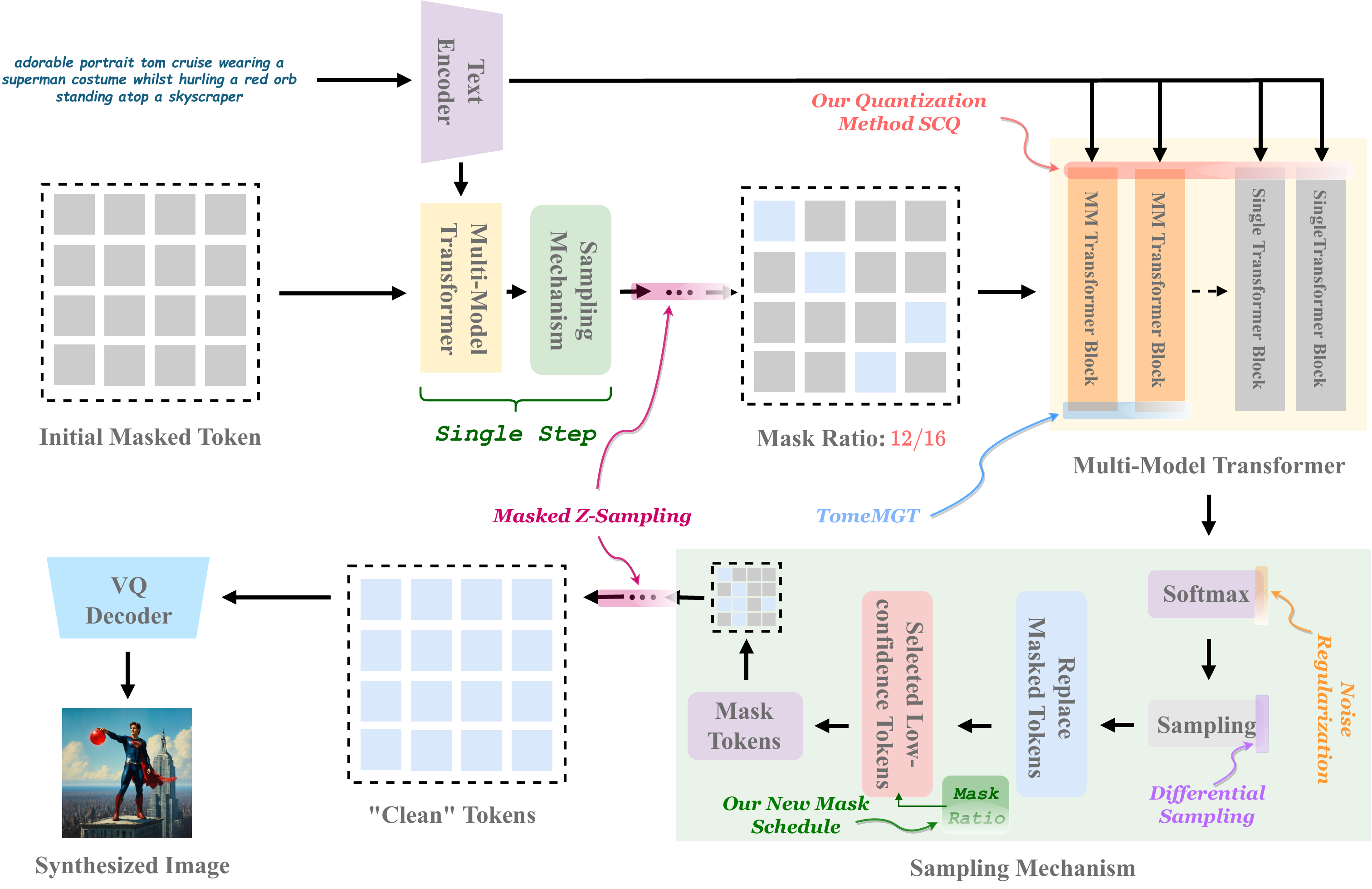}
\vspace{-6pt}
\caption{The complete sampling pipeline of MaskGIT and Meissonic, and how our proposed effective and specific design choices integrate into that sampling pipeline. Specifically, TomeMGT and SCQ act on the Transformer to reduce inference latency and memory usage, respectively. Meanwhile, noise regularization and differential sampling enhance inference by correcting the probability distribution applied for sampling. Additionally, masked Z-Sampling is a rescheduling technique that significantly improves the quality of synthesized images through the forward-inversion operator (\textit{i.e.}, sampling and backtracking alternatively).}
\vspace{-12pt}
\label{figure:total_framework}
\end{figure*}

To fill this gap, this paper focuses on the inference phase of MGT, with the primary objective of identifying design choices that enhance visual quality and the secondary goal of achieving efficient sampling through empirical analysis in high-resolution image synthesis scenarios. \textcolor{C3}{\textit{First}}, we elucidate how well-known training-free methods in DM can be applied to MGT, and the redesign required to ensure their effectiveness. As illustrated in Fig.~\ref{figure:total_framework} and Table~\ref{tab:specifics}, the sampling process of MGT bears significant similarity to that of DM, making it intuitive to adapt those algorithms from DM to MGT. We explore DPM-Solver~\citep{dpm_solver,dpm_solver++}, TomeSD~\citep{tomesd}, and Z-Sampling~\citep{zigzag} in this context, but find that all three algorithms require specific modifications to align with the characteristics of MGT in order to reduce NFE, accelerate inference, or achieve performance improvements. Given an example of Z-Sampling, we find that implementing DDIM Inversion~\citep{ddiminversion}, as used in DM, with random masking is ineffective unless masking is limited to low-confidence predicted tokens. In particular, our experimental outcomes indicate that among these three algorithms, DPM-Solver and TomeSD have relatively limited effects on MGT, whereas a rescheduling algorithm like Z-Sampling yields notable performance gains. Furthermore, we investigate the noise schedule in MGT, akin in EDM~\citep{karras2022elucidating}, highlighting that the cosine noise schedule is suboptimal under certain conditions. These findings suggest that inconsistencies between the training and inference mechanisms of DM and MGT lead to enhanced inference algorithms on DM not necessarily being effective for MGT.


\textcolor{C3}{\textit{Second}}, we take a look at the probability distribution generated by the Transformer, leading to the development of several ``cheap'' (\textit{i.e.}, w/o significant computational overhead) yet effective distribution correction algorithms, including noise regularization and (low-entropy) differential sampling. To be specific, noise regularization dynamically applies (Gaussian) perturbations to the backbone output based on timesteps before applying \textit{softmax}, aiming to enhance the diversity of synthesized images. Differential sampling, on the other hand, calculates the Kullback-Leibler (KL) divergence between the outputs of two adjacent time steps and resamples tokens with excessively similar Transformer outputs, thereby avoiding information redundancy and enhancing the visual quality.

\textcolor{C3}{\textit{Third}}, we also investigate the model quantization on Meissonic for efficient memory usage. Our results reveal that Weight4Activation16 (W4A16) quantization fails to reduce memory usage in practice, while W4A8 quantization results in inference collapse. To address this issue, we quantize only the layers with low-amplitude activation values, reducing the memory footprint from 11.98 GB to 4.57 GB w/o significant performance degradation.

\input{tables/tabspecific}
\textcolor{C3}{\textit{Fourth}}, we evaluate the effectiveness of our proposed design choices on Meissonic using the HPD v2 benchmark, employing various metrics such as ImageReward~\citep{Imagereward}, HPS v2~\citep{HPSV2}, PickScore~\citep{fid,COCO}, and AES~\citep{AES}. Similarly, we assess the performance of our strategy on MaskGIT using traditional metrics, including IS~\citep{is} and FID~\citep{fid}. As shown in Fig.~\ref{figure:visualization}, these strategies lead to a significant improvement in the visual quality of synthesized images from Meissonic-512$\times$512 and Meissonic-1024$\times$1024.




%% file: tables/tabspecific.tex
\tabulinesep=0.00ex%
\tabulinestyle{0.17mm}%
\begin{table*}[t]%
\centering%
\caption{\label{tab:specifics}%
Specific design choices employed by masked generative Transformers (MGTs) are presented in this overview. We adopt a definitional form of sampling that is consistent with DMs, akin to EDM~\cite{karras2022elucidating}. Let $N$ denote the number of sampling steps, and the sequence of time steps is $\{t_0,\cdots,t_N\}$, where $\sigma_{t_N}=0$. Furthermore, the strategies highlighted in \textcolor{C1}{yellow} are our proposed methods.}
\vspace{1.5mm}%
\resizebox{\textwidth}{!}{%
\begin{tabular}{@{}p{3.6cm}cccc@{\hspace*{-1mm}}}
\toprule
& {\bf DM~\cite{sde}}
& {\bf ARM~\cite{llamagen}}
& {\bf MGT~\cite{meissonic}}
& {\bf Ours}
\\
\hline\vspace*{-1.5mm}
\titlerowww{Definition (Section~\protect\ref{sec:related_work})}
TimeStep\hfill$t_{0\leq i\leq N}$ & \makecell[l]{$t=1+\frac{i}{N}(\epsilon-1)$\\\quad\quad(VP-SDE \& flow matching)\\$i/N$\hfill(EDM)}
& {\makecell[c]{N/A\\(next-token prediction)}}
& {\makecell[c]{$i/N$\\(non-ar token prediction)}}
& {\makecell[c]{$i/N$\\(non-ar token prediction)}} \vspace*{+1.2ex}\\
\makecell[l]{Noise Schedule\\(Section~\ref{sec:convex_exploration})}\hfill$\sigma_t$
& \makecell[l]{$\sqrt{e^{at^2+bt}-1}$\hfill(VP-SDE~\citep{sde})\\$t$\hfill(flow matching~\citep{iclr22_rect})\\$(\sigma_\textrm{max}^{\frac{1}{\rho}}+t(\sigma_\textrm{min}^{\frac{1}{\rho}}-\sigma_\textrm{max}^{\frac{1}{\rho}}))^\rho$\hfill(EDM~\citep{karras2022elucidating})}
& {\makecell{N/A, and predicts\\one token per iteration}}
& {$\mathrm{cos}\left(\frac{\pi t}{2}\right)$}
& {\makecell{\textcolor{C1}{$(1-t)^\rho$ or $1-t^\rho$}}} \vspace*{+1.2ex}\\
\makecell[l]{\makecell{Network\\Architecture}}\hfill$f_\theta$ & \makecell[l]{U-Net~\citep{ddpm_begin} or Transformer~\citep{DIT}\\(encoder only)} & \makecell{Transformer~\citep{llamagen}\\(decoder only)} &  \makecell{Transformer~\citep{meissonic}\\(encoder only)} &  \makecell{Transformer~\citep{meissonic}\\(encoder only)}
\vspace*{+1.2ex}\\
\makecell[l]{\makecell{Coding Form}}\hfill$Q(\ve{z}|\ve{x})$ & \makecell[l]{VAE~\citep{VAE} (continuous)} & \makecell{VQ-VAE~\citep{VQVAE} (discrete)} &  \makecell{VQ-VAE~\citep{VQVAE} (discrete)} &  \makecell{VQ-VAE~\citep{VQVAE} (discrete)}
\vspace*{-0.2mm}\\
\hline\vspace*{-1.5mm}
\titlerowww{Enhanced Inference (Section~\protect\ref{sec:strong_inference})}
\makecell[l]{Sampling Paradigm\\$p(\ve{z}_i|\prod_{j<i}\ve{z}_{j})$\ \hfill(Sec.~\ref{sec:sampling paradigm})}& \makecell[c]{DDPM~\citep{ddpm_begin}, Euler Maruyama~\citep{sde},\\Classifier-free Guidance~\citep{nips2021_classifier_free_guidance},\\Z-Sampling~\citep{zigzag}, et al.} & \makecell{Autoregressive\\($\ve{z}_i$ denotes a token)} & \makecell{MaskGIT's Sampling~\citep{maskgit}\\($\ve{z}_i$ denotes all masked tokens)}  & \makecell{\textcolor{C1}{Masked Z-Sampling, \textit{i.e.}}\\\textcolor{C1}{$p(\ve{z}_i|\tilde{\ve{z}}_i,\prod_{j<i}\ve{z}_{j})$, where $\tilde{\ve{z}}_i\sim p(\ve{z}_i|\prod_{j<i}\ve{z}_{j})$}\\\textcolor{C1}{($\ve{z}_i$ denotes all masked tokens)}}\\
\makecell[l]{Improved Probability\\Distribution (Sec.~\ref{sec:noise_regularization}\ \&\ \ref{sec:differential_pro})}& N/A & \makecell{$\arg\max_i \frac{\log(\epsilon)}{\ve{p}}$,\\where $\ve{p}$ is the logit\\and $\epsilon\sim \mathcal{U}[\ve{0},\ve{1}]$} & \makecell{$\arg\max_i \frac{\log(\epsilon)}{\ve{p}}$,\\where $\ve{p}$ is the logit\\and $\epsilon\sim \mathcal{U}[\ve{0},\ve{1}]$} & \makecell{\textcolor{C1}{Noise regularization and}\\\textcolor{C1}{differential sampling}}\\
\hline\vspace*{-1.5mm}\titlerowww{Efficient Inference (Section~\ref{sec:accelerated_inference})}\makecell[l]{Quantization (Section~\ref{sec:model_quantization})} & \makecell{Both Int4 and Int8\\have been successfully~\citep{QDIT}} & N/A & N/A & \makecell{\textcolor{C1}{W4A8 Quantization} \textcolor{C1}{(Our proposed SCQ)}} \\
\makecell[l]{Token Merging (Section~\ref{sec:token_merging})} & \makecell{TomeSD~\citep{tomesd}} & N/A & N/A & \makecell{\textcolor{C1}{TomeMGT}\\\textcolor{C1}{(Transfer TomeSD into MGT)}} \\
\makecell[l]{Reduce NFE (Section~\ref{sec:momentum_based_solver})} & \makecell{DDIM~\citep{ddim}, Deis~\citep{zhang2023fast}\\and DPM-Solver~\citep{dpm_solver}} & Jacobi Decoding~\citep{teng2024accelerating} & N/A & \makecell{\textcolor{C1}{Momentum-based Solver}} \vspace*{-3mm}\\\bottomrule
\multicolumn{5}{l@{}}{\hfill%
W4A8: the weights are quantized to 4 bits and the activation values are quantized to 8 bits.\hfill%
$a$, $b$, $\sigma_\textrm{max}$, $\sigma_\textrm{min}$: 9.95, 0.1, 80, 0.002.\hfill~}
\end{tabular}}
\vspace{-14pt}
\end{table*}

%% file: sec/related_work.tex
\vspace{-5pt}
\section{Preliminaries}
\label{sec:related_work}

We begin by reviewing three generative models that are experiencing continuous growth in the field of vision synthesis: diffusion model (DM)~\citep{ddpm_begin}, autoregressive model (ARM)~\citep{llamagen}, and masked generative Transformer (MGT)~\citep{meissonic}. We then provide an overview of MGT's vanilla sampling process, which is introduced by MaskGIT~\citep{maskgit}.
\vspace{-6pt}
\paragraph{Inference Mechanism of Visual Generative Model.} DM is a well-established technique that has developed in recent years and has successfully scaled to large-scale, high-quality visual synthesis, whereas ARM~\citep{luminia_mgpt} and MGT~\citep{meissonic} have only recently demonstrated feasibility for synthesizing high-resolution images. As illustrated in Table~\ref{tab:specifics}, each of them employs the multi-step denoising paradigm to progressively generate high-quality images during inference. Given a latent variable $ \ve{z}_{t_0} $ ($t$ is defined in Table~\ref{tab:specifics}), it may follow a Gaussian distribution (\textit{w.r.t.}, DM) or consist of masked tokens (\textit{w.r.t.}, ARM and MGT). These methods primarily expect to fit the (abstract) estimator $p(\ve{z}_{t_i}|\ve{z}_{t_{i-1}})$ ($i\geq 1$) in the training phase, thus enabling the sequential sampling $p(\ve{z}_{t_N}|\prod_{i=0}^{N-1}\ve{z}_{t_{i}})=p(\ve{z}_{t_0})\prod_{i=1}^{N}p(\ve{z}_{t_i}|\ve{z}_{t_{i-1}})$ during inference, where $N$ stands for the number of sampling steps. Note that $p(\ve{z}_{t_i}|\ve{z}_{t_{i-1}})$ can be instantiated in various models with distinct focuses: in DM, it represents the prediction of a score function; in ARM, it embodies the prediction of a token; and in MGT, it pertains to the prediction of all masked tokens. In both DM and MGT, the encoder-only Transformer is employed to predict the complete score function or the full tokens, allowing their sampling process to be refined as $p(\ve{z}_{t_i}|\ve{z}_{t_{i-1}})=\int_{\ve{z}_{t_N}} p(\ve{z}_{t_i}|\ve{z}_{t_{i-1}},\ve{z}_{t_N})p(\ve{z}_{t_N}|\ve{z}_{t_{i-1}})d\ve{z}_{t_N}$.
\vspace{-6pt}
\paragraph{Vanilla Sampling Process of MGT.} The sampling process for MGT was given by MaskGIT~\citep{maskgit} and subsequently followed by Muse~\citep{muse} and Meissonic~\citep{meissonic}. As illustrated in Fig.~\ref{figure:total_framework}, the complete sampling process of MGT closely resembles counterpart of DM, though they differ in several critical aspects. To be specific, given the initial masked tokens $\ve{z}_{t_0}$, it is sampled in multiple steps to obtain the ``clean'' tokens $\ve{z}_{t_N}$. Taking Meissonic as an example, each step $i$ involves: \textcolor{C3}{\textit{1)}} the MM Transformer giving the predicted tokens; \textcolor{C3}{\textit{2)}} replacing the masked tokens in $\ve{z}_{t_{i}}$ with the predicted tokens, followed by masking out tokens with low confidence based on their probability values. The main differences between the sampling processes of MGT and DM are: \textcolor{C3}{\textit{1)}} each step of MGT is modeled non-deterministically (see Table~\ref{tab:specifics}), and forcing deterministic sampling is likely to degrade performance (\textcolor{black}{see Appendix~\ref{apd:deterministic_sampling} for details}); \textcolor{C3}{\textit{2)}} in MGT, the predicted tokens only replace the masked tokens and do not affect the unmasked tokens; and \textcolor{C3}{\textit{3)}} how the masking is performed in MGT is determined by probability values, unlike DM, which is random.


%% file: sec/method.tex
\vspace{-5pt}
\section{Enhanced Inference}
\label{sec:strong_inference}

Sec.~\ref{sec:strong_inference} and Sec.~\ref{sec:accelerated_inference} will discuss our explorations of enhanced and efficient inference, respectively. In this section, enhanced inference involves adaptations of well-known DM methods, along with new algorithms designed based on the properties of MGT. Note that to introduce our research more logically, we will introduce our methods and experiments in the form of a progressive exploration. For follow-up content, we employ the definitions in Table~\ref{tab:specifics}. All experiments, unless otherwise specified, were conducted with Meissonic-1024$\times$1024 on the HPD v2 Subset (see Appendix~\ref{apd:benchmark}).

\subsection{Convexity Exploration of Noise Schedule}
\label{sec:convex_exploration}

\begin{figure}[h]
\vspace{-8pt}
\centering
\hspace{-0.5cm}
\includegraphics[width=0.5\textwidth,trim={0cm 0.0cm 0cm 0cm},clip]{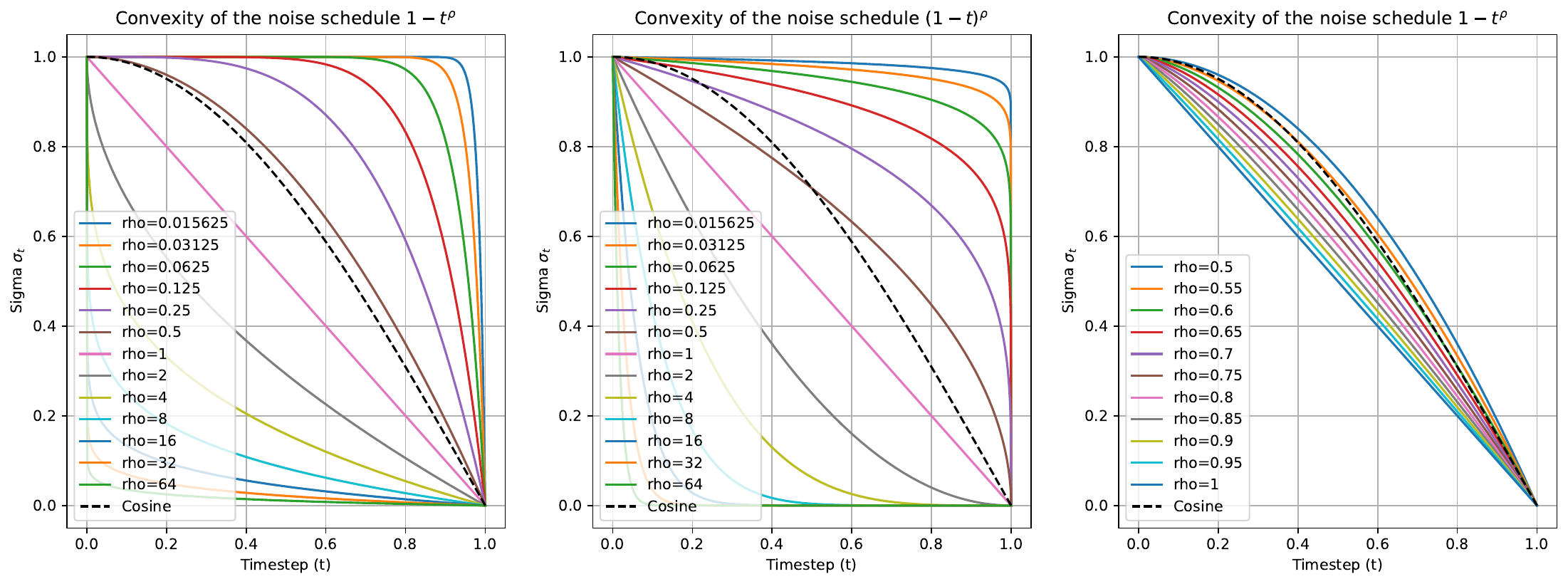}
\vspace{-20pt}
\caption{Visualization of different noise schedules. The black dashed line represents the cosine schedule.}
\vspace{-8pt}
\label{figure:karras_sigma}
\end{figure}

\begin{figure}[h]
\vspace{-8pt}
\centering
\hspace{-0.5cm}
\includegraphics[width=0.5\textwidth,trim={0cm 0.0cm 0cm 0cm},clip]{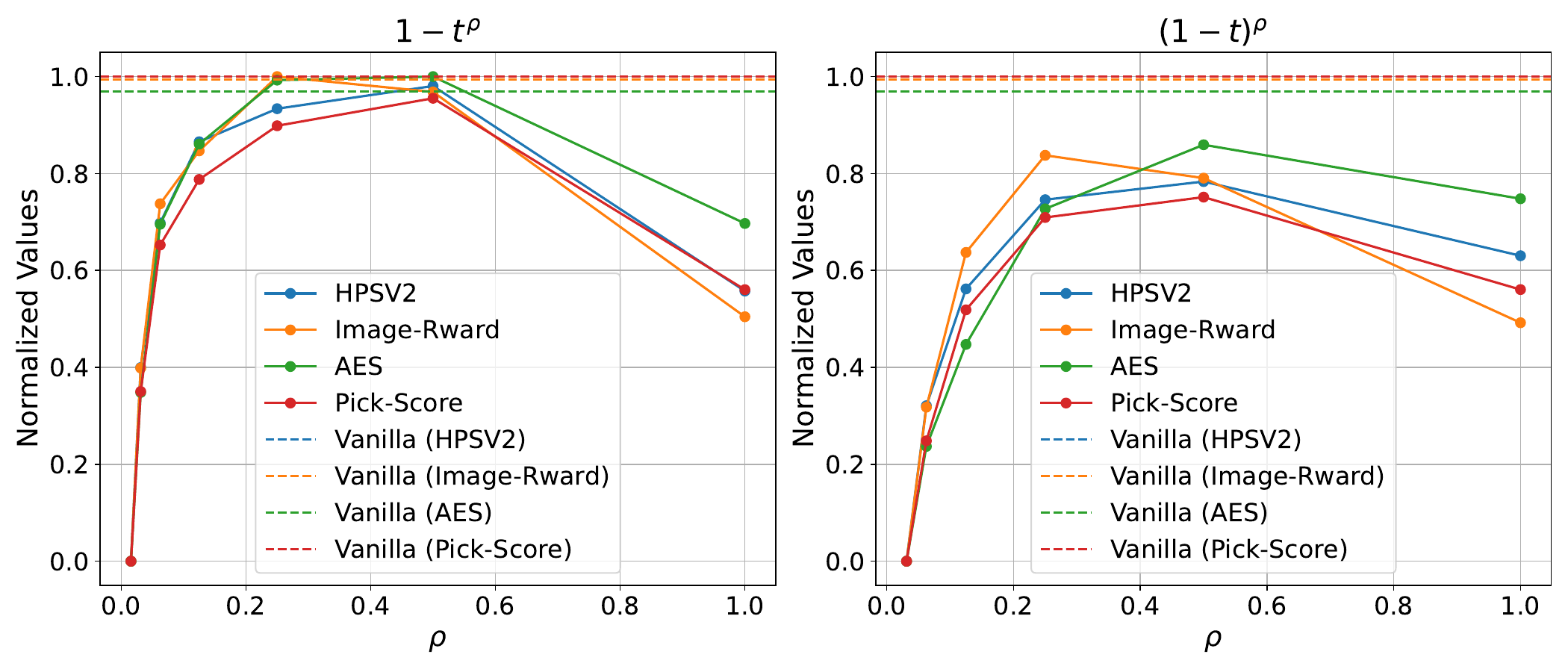}
\vspace{-20pt}
\caption{Visualization of the performance of different noise schedules. The dotted line denotes the vanilla sampling.}
\vspace{-8pt}
\label{figure:karras_sigma_result}
\end{figure}

All known MGT models use the cosine schedule shown in Table~\ref{tab:specifics} for both training and inference to ensure consistency. To investigate whether a better noise schedule exists for inference, we explore curves with different convexities (see Fig.~\ref{figure:karras_sigma}). The curve design is inspired by Karras' noise schedule~\citep{karras2022elucidating}. Specifically, the expressions $(\sigma_\textrm{begin}^{\frac{1}{\rho}} + t(\sigma_\textrm{end}^{\frac{1}{\rho}} - \sigma_\textrm{begin}^{\frac{1}{\rho}}))^\rho$ and $1-((1-\sigma_\textrm{begin})^{\frac{1}{\rho}} + t((1-\sigma_\textrm{end})^{\frac{1}{\rho}} - (1-\sigma_\textrm{begin})^{\frac{1}{\rho}}))^\rho$ can be simplified to $(1-t)^\rho$ and $1-t^\rho$ when $\sigma_\textrm{begin}$ is set to $1$ and $\sigma_\textrm{end}$ to $0$. We present the experimental results in Fig.~\ref{figure:karras_sigma_result} and Table~\ref{tab:karras_sigma_01}, where the number of sampling steps $N$ is set to $64$.

The key takeaway from Fig.~\ref{figure:karras_sigma_result} is that $1-t^\rho$ outperforms $(1-t)^\rho$, with the most metrics peaking around $\rho = 0.5$. Notably, some metrics even exceed the performance of vanilla sampling. Given this observation, we sample $\rho$ more densely and uniformly within the interval $[0.5,1]$ to obtain more precise results, which is presented in Table~\ref{tab:karras_sigma_01}.\input{tables/karras_inv_sigma_01}

When $\rho=0.6$, $1-t^\rho$ demonstrates more favorable behavior compared to the standard cosine schedule $\mathrm{cos}(\frac{\pi t}{2})$. This substantiates that, even if the cosine schedule is used during training, a better noise schedule may exist for inference. We also present additional experimental results for different values of $N$ and benchmarks in Appendix~\ref{sec:additional_experiment_noise_schedule}.

\subsection{Masked Z-Sampling for MGT}
\label{sec:sampling paradigm}
\begin{figure}[h]
\vspace{-8pt}
\centering
\hspace{-1pt}
\includegraphics[width=0.5\textwidth,trim={0cm 0.0cm 0cm 0cm},clip]{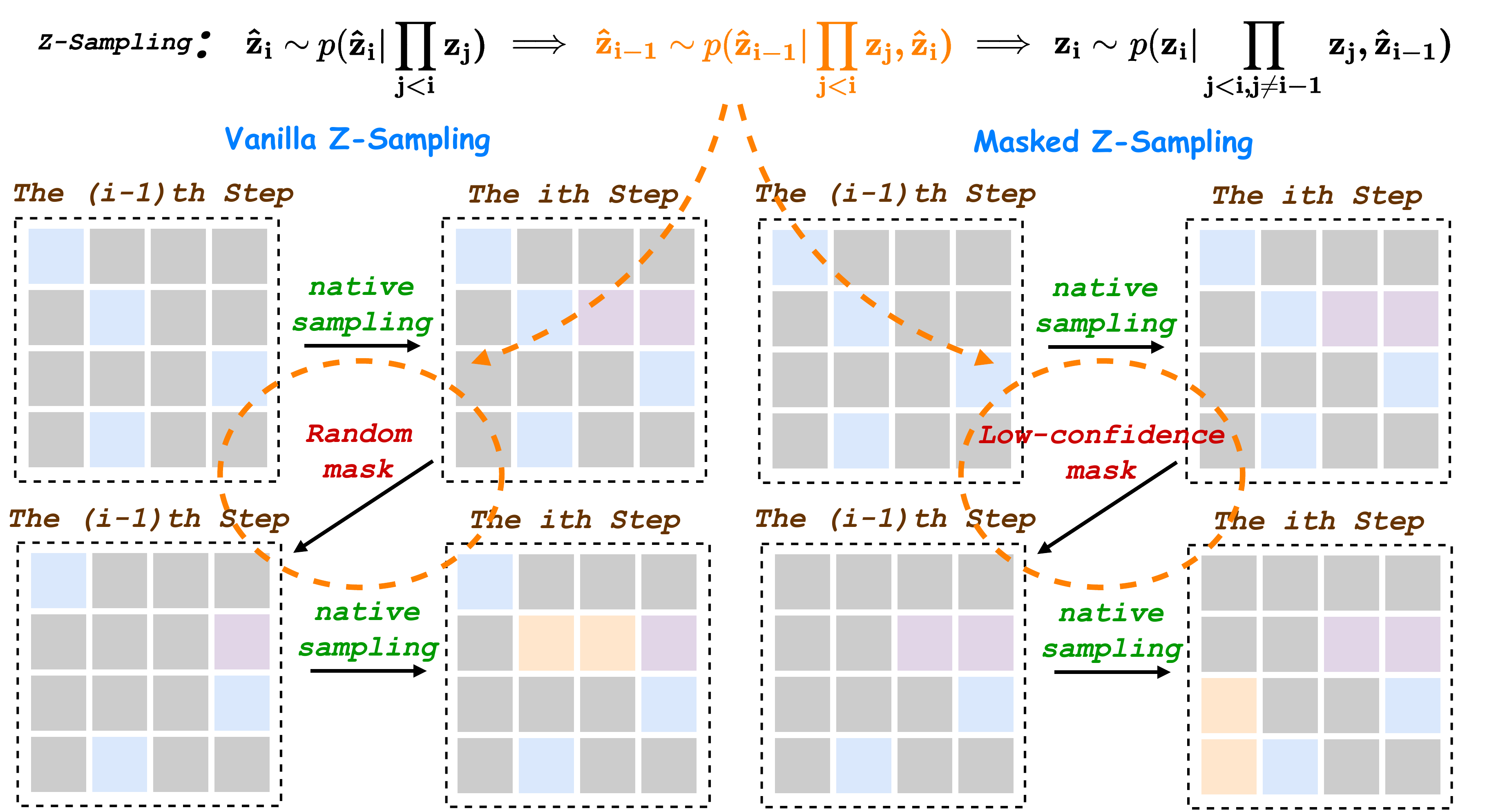}
\vspace{-20pt}
\caption{The illustration of vanilla Z-Sampling and masked Z-Sampling. The main difference is the masking form.}
\vspace{-8pt}
\label{figure:zigzag_sampling_illustration}
\end{figure}

The core idea of Zigzag diffusion sampling (Z-Sampling)~\citep{cvpr22_kd_guided,zigzag} is to improve the sampling quality of DMs by incorporating ``future'' semantic information in advance, using a ``zigzag'' path for sampling. We aim to extend this algorithm, which has demonstrated effective for DMs, to MGT to enhance the fidelity of synthesized images. The logic of Z-Sampling is illustrated by the equation at the top of Fig.~\ref{figure:zigzag_sampling_illustration}. After obtaining the latent $\ve{\hat{z}}_i$, it backtracks to $t$=i$-$1 using the ``specific'' masking algorithm (corresponds to DDIM Inversion in DMs) and performs sampling from $t$=i$-$1 to $t$=i again.

Unfortunately, applying random masking (\textit{i.e.}, vanilla Z-Sampling in Fig.~\ref{figure:zigzag_sampling_illustration}) to simulate DDIM inversion in DMs impaired inference performance in our experiments. We argue that this is due to random masking incorrectly removing certain tokens in the latent space that significantly contribute to the synthesized image. For instance, the purple token obtained during the $1$st forward sampling in Fig.~\ref{figure:zigzag_sampling_illustration} may be masked out, even though these purple tokens typically represent the most ``future'' information. Therefore, we employ a novel masking pipeline for backtracking that is consistent with that of sampling mechanism, specifically masking the portion of the predicted token at the $i$th step with the low log probability (\textit{i.e.}, masked Z-Sampling in Fig.~\ref{figure:zigzag_sampling_illustration}). We also need to mention an important parameter: the inversion classifier-free guidance (CFG) scale, which refers to the CFG scale used to generate tokens for selecting low-confidence probabilities during the masking phase. We looked at how the inversion CFG scale affect the quality of synthesized images. As outlined in~\citep{zigzag} that a ``just right'' CFG gap can be produced by selecting a desirable inversion CFG scale that maximizes the positive impact of semantic information injection.

\begin{figure}[h]
\vspace{-8pt}
\centering
\hspace{-1pt}
\includegraphics[width=0.5\textwidth,trim={0cm 0.0cm 0cm 0cm},clip]{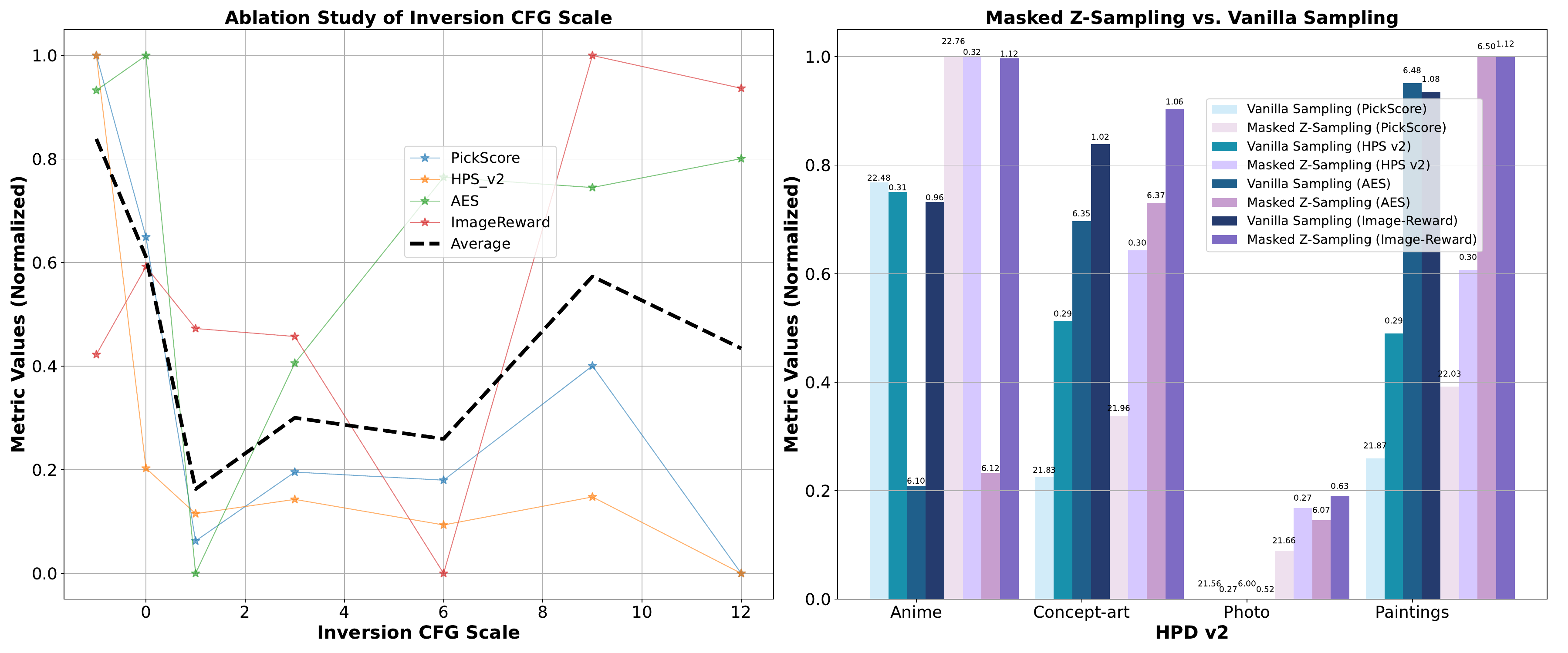}
\vspace{-20pt}
\caption{\textbf{Left:} Ablation study on the inversion CFG scale. \textbf{Right:} Comparison between masked Z-Sampling and vanilla sampling on the HPD v2 dataset.}
\vspace{-8pt}
\label{figure:zigzag_sampling_ablat_compar}
\end{figure}

We present the ablation results in Fig.~\ref{figure:zigzag_sampling_ablat_compar} (Left). From the change in the black dashed line (\textit{i.e.}, average metric), it can be concluded that the inversion CFG scale performs best near -1 and 9. To reduce computational cost, we set the inversion CFG scale to 0 (\textit{i.e.}, do not use CFG) and the standard CFG scale to 9 in our experiments, thereby avoiding additional computational overhead by reducing the NFE.

We further validate masked Z-Sampling on the HPD v2 dataset, with the results presented on Fig.~\ref{figure:zigzag_sampling_ablat_compar} (Right). It can be observed that our algorithm significantly outperforms vanilla sampling across nearly all domains and metrics, demonstrates that masked Z-Sampling can steadily enhance the performance of MGT.

\subsection{Noise Regularization}
\label{sec:noise_regularization}

According to our research, a major difference between MGT and DMs is that MGT can improve the visual quality and diversity of synthesized images by adjusting the model outputs' probability distribution. As a result, noise regularization and differential sampling are proposed. Here, we first present a simple yet effective approach known as noise regularization, which can be described as\begin{equation}
    \begin{split}
        & \ve{v}_i = f_\theta(\ve{z}_{i},t_{i}), \\
        & \textcolor{C1}{\ve{\hat{v}}_i =  \ve{v}_i  + \epsilon_t,\textrm{ where }\epsilon_t\sim \mathcal{N}(\mathbf{0},\mathbf{I}_t),} \\
        & \ve{p}_i = \mathrm{softmax}(\ve{\hat{v}}_i),\\
    \end{split}
\end{equation}where $\ve{v}_i$ is the output of $f_\theta$ and the \textcolor{C1}{yellow} part stands for noise regularization. Noise regularization introduces randomness (see Appendix~\ref{apd:analysis_noise_reg} for more details) into the sampling process, enhancing the diversity of predicted tokens.\begin{figure}[h]
\vspace{-8pt}
\centering
\hspace{-0.5cm}
\includegraphics[width=0.5\textwidth,trim={0cm 0.0cm 0cm 0cm},clip]{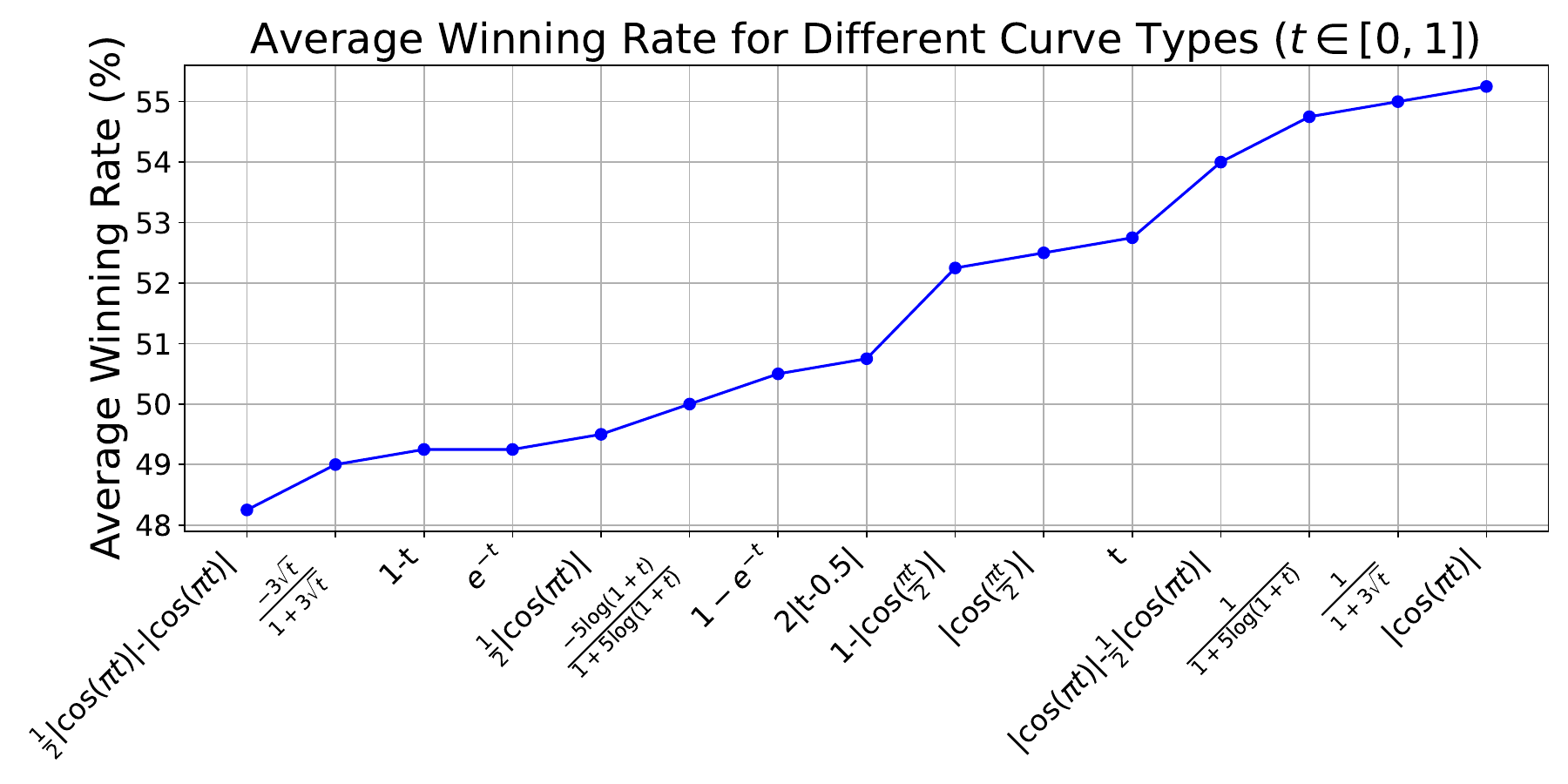}
\vspace{-20pt}
\caption{Ablation study on \(\mathbf{I}(t)\). Note that \(\frac{1}{2}|\cos(\pi t)|\!-\!|\cos(\pi t)|\) and \(|\cos(\pi t)|\!-\!\frac{1}{2}|\cos(\pi t)|\) represent the scenarios where the function \(\frac{1}{2}|\cos(\pi t)|\) or \(|\cos(\pi t)|\) is used for \(t \in [0, \frac{1}{2})\), while \(|\cos(\pi t)|\) or \(\frac{1}{2}|\cos(\pi t)|\) is used for \(t \in [\frac{1}{2}, 1]\), respectively.}
\vspace{-8pt}
\label{figure:noise_regularization_winning_rate}
\end{figure}

In our definition, the hyperparameter $\mathbf{I}_t$ is a function $\mathbf{I}(t)$ that represent the standard deviation of the noise, which varies across different timesteps $t$. To determine the empirically optimal $\mathbf{I}(t)$, we test various curves and calculate the winnowing rates of four metrics—PickScore, HPS v2, AES, and ImageReward—relative to vanilla sampling. The results are presented in Fig.~\ref{figure:noise_regularization_winning_rate}. We find that $\mathbf{I}(t)$ works best when set to $|\cos(\pi t)|$, which reaches larger values as $t$ approaches $0$ or $1$ and lower values in the middle. This observation is interesting, as injecting noise during the end-of-sampling phase significantly degrades inference performance on DM, while having positive impact on MGT. The results on HPD v2 can be found in Appendix~\ref{apd:comparison_noise_regularization}.


\subsection{Differential Sampling}
\label{sec:differential_pro}
\begin{figure}[h]
\vspace{-8pt}
\centering
\hspace{-0.5cm}
\includegraphics[width=0.5\textwidth,trim={0cm 0.0cm 0cm 0cm},clip]{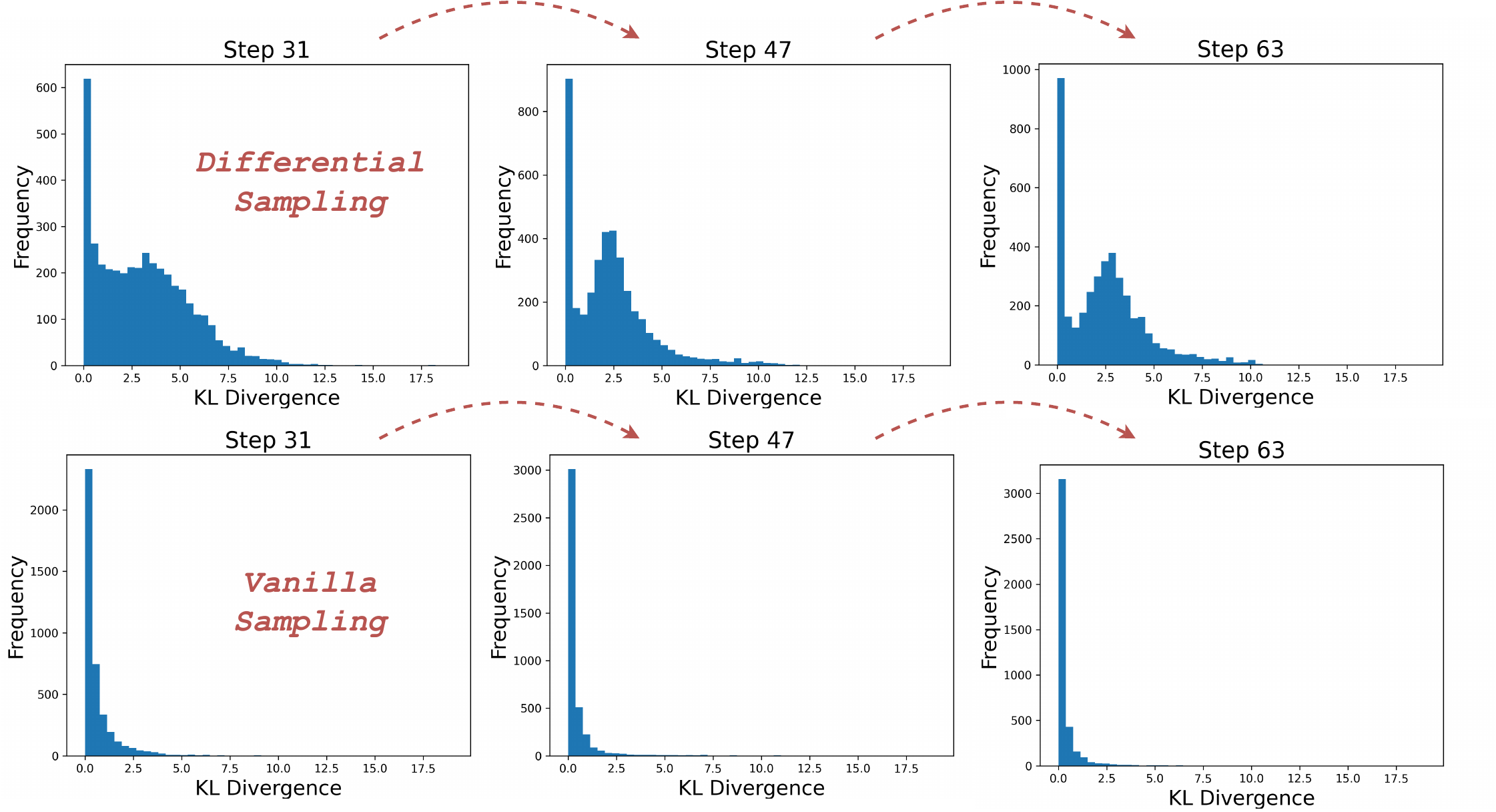}
\vspace{-20pt}
\caption{Trajectory of KL divergence between distributions at neighboring time steps throughout the sampling process. Note that the variability in the distributions of two consecutive steps in vanilla sampling is minimal as KL divergence are all close to 0.}
\vspace{-14pt}
\label{figure:differential_sampling_kl}
\end{figure}
The effectiveness of noise regularization highlights the importance of enhancing diversity in MGT by adjusting the probability distribution. In addition to this, we show in Fig.~\ref{figure:differential_sampling_kl} (Bottom) that the probability distribution tends to overly rely on the outputs from certain timesteps, which leads to the propagation of similarity throughout the sampling process. In response to this issue, we propose (low-entropy) differential sampling, which resamples the probability distribution of the current step when it is too similar to that of the previous step. Assume that the probability distribution of the $j$th token at the $i$th step is denoted by $\ve{p}_i^j$ ($i \geq 1$). The KL divergence between distributions at the $i$th step and the $(i\!-\!1)$th step can then be expressed as
\begin{equation}
    \begin{split}
        & \mathcal{D}_i = \{d^1_i, d^2_i,\cdots, d^K_i\},\quad d^j_i = \mathcal{D}_\mathrm{KL}(\ve{p}_i^j\Vert \ve{p}_{i-1}^j),\\
    \end{split}
    \label{eq:kl_divergence}
\end{equation}
where $\mathcal{D}_i$, $K$ and $\mathcal{D}_\mathrm{KL}(\cdot \Vert \cdot)$ refer to the KL divergence set, the number of tokens, KL divergence, respectively. To identify the tokens with the worst propagation of similarity, we sort $\mathcal{D}_i$ and reject the lowest z\% of tokens and then resampling them, where the resampling comes from the differential between two probability distributions:\begin{equation}
    \begin{split}
        & \ve{\tilde{p}}_i^j = \frac{|\ve{p}_i^j - \ve{p}_{i-1}^j|}{\sum_{p \in |\ve{p}_i^j - \ve{p}_{i-1}^j|} p}.\\
    \end{split}
    \label{eq:kl_divergence_new}
\end{equation}
\begin{figure}[!h]
\vspace{-16pt}
\centering
\hspace{-0.5cm}
\includegraphics[width=0.5\textwidth,trim={0cm 0.0cm 0cm 0cm},clip]{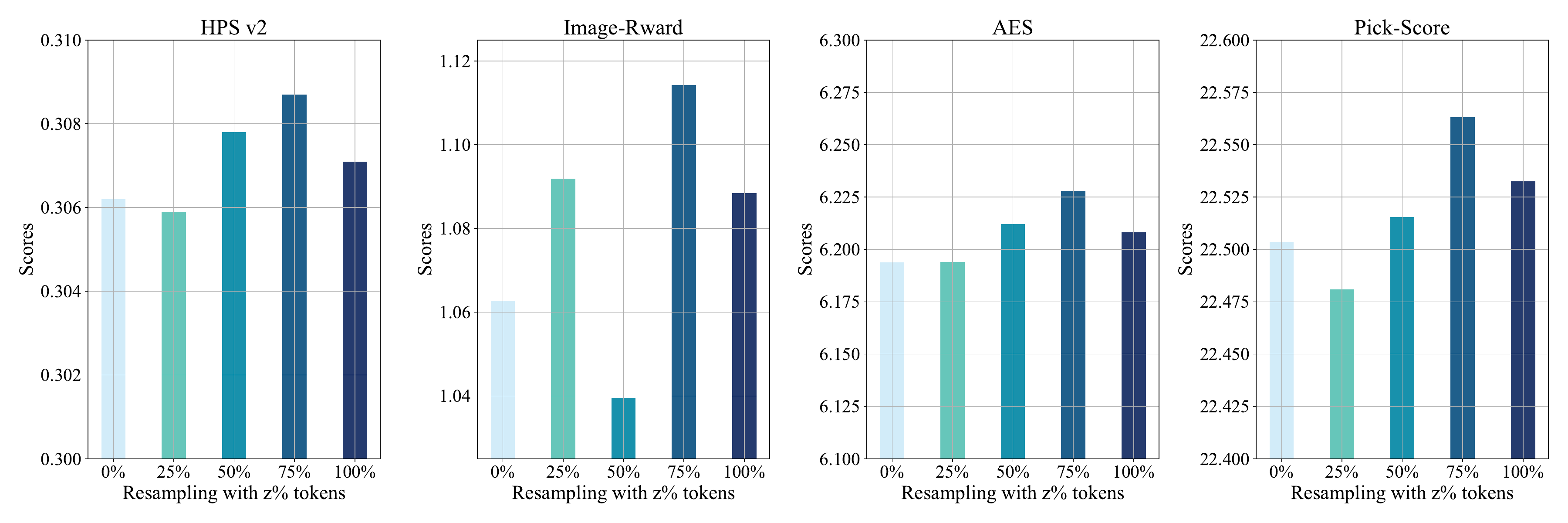}
\vspace{-20pt}
\caption{Ablation Studies of $z$\% in differential sampling.}
\vspace{-15pt}
\label{figure:differential_sampling_ablation_z}
\end{figure}
\input{tables/model_quantization_ablation_study}
\input{tables/challengebench}
As shown in Fig.~\ref{figure:differential_sampling_kl} (Top), this approach can significantly mitigate the propagation of similarity and effectively introduce diversity into the sampling process. Moreover, we conduct ablation experiments on the hyperparameter $z$, and the results are presented in Fig.~\ref{figure:differential_sampling_ablation_z}. As $z$ increases from 0 to 100~($z$=0 means vanilla sampling), the sampling performance initially improves and then declines. Interestingly, differential sampling outperforms vanilla sampling even when applied to all tokens, highlighting the robustness of differential sampling. Furthermore, empirical results indicate that the best performance is achieved when $z$ is set to 75., and we present the performance of differential sampling on HPD v2 in Appendix~\ref{apd:comparison_differential_sampling}.

\section{Efficient Inference}
\label{sec:accelerated_inference}

Another path we explore is determining how to achieve efficient inference on MGT. We consider model quantization~\citep{jacob2018quantization}, token merging~\citep{tomesd}, and scheduling strategies similar to those used in DM~\citep{dpm_solver,dpm_solver++,qi2024not}.

\subsection{Secondary Calibration for Quantization}
\label{sec:model_quantization}

The most effective way to achieve memory efficiency is to apply model quantization to the backbone of generative model, a technique successfully used in FLUX~\citep{FLUX} and SD 3.5~\citep{SD35}. Unfortunately, this approach does not work on Meissonic-1024$\times$1024 due to \textcolor{C3}{\textit{1)}} its limited number of model parameters (only 1 billion) and its compression layer that actively reduces the number of tokens to 1024. These constraints lead to issues when applying W4A16 post-training quantization (PTQ), resulting in an inability to synthesize normal images. In addition, \textcolor{C3}{\textit{2)}} since Meissonic incorporates a multi-model Transformer block, the overly complex architectural design prevents the quantized memory from being significantly reduced in practice. A straightforward solution is to quantize the activation values. However, this operation further degrades model performance. 

To address these issues, we propose secondary calibration for quantization (SCQ). Our core contribution involves \textcolor{C3}{\textit{1)}} performing quantization-aware training (QAT) using Meissonic's synthesized images to correct the range of quantized values, followed by \textcolor{C3}{\textit{2)}} introducing a secondary calibration strategy that records the magnitude of each layer after primary calibration and subsequently quantizes only the activation values with smaller magnitudes, further calibrating them. In our experiments, we quantized only one-third of the activation values by default, which reduced the backbone's memory usage from 3.34 GB to 2.24 GB.

The experimental results of SCQ are presented in Table~\ref{tab:memory_comparison}, where ``A8W4-QAT \& Calibration'' represents a single calibration step performed on the Transformer (i.e., backbone) derived from QAT. For a fair comparison, one-third of the activation values were randomly selected for quantization in ``A8W4-QAT \& Calibration''. Additionally, ``CPU offloading'' indicates that the Transformer was used first to obtain all $\ve{z}_N$, after which the tokenizer decoder is loaded to transform $\ve{z}_N$ into synthesized images. From Table~\ref{tab:memory_comparison}, it can be concluded that both QAT and secondary calibration strategies are critical and effective.

\subsection{Introducing TomeSD into MGT}
\label{sec:token_merging}

Token merging is designed to enhance efficiency by first merging tokens after the linear layers and subsequently unmerging them. Applying token merging to accelerate inference is natural, as the backbone of MGT is a Transformer. Unfortunately, Meissonic has only 1024 tokens, fewer than 4096 tokens in the attention layer of SD XL. As is well known, the computational complexity of attention increases exponentially with the number of tokens, and a smaller number of tokens reduces the potential benefits of token merging, resulting in an insignificant effect of our implemented TomeMGT as observed in our experiments. Therefore, we focus on the challenges of applying TomeSD, which has demonstrated effective on SD XL, to MGT to achieve accelerated inference, along with the corresponding application scenarios.

The main challenge consists of two aspects. \textcolor{C3}{\textit{First}}, applying token merging to a single Transformer may cause inference to fail, whereas it is effective in a multi-modal Transformer. \textcolor{C3}{\textit{Second}}, RoPE~\citep{RoPE} (used for encoding positional information) in Meissonic also requires merging. For the former, we perform token merging only on multi-modal Transformers, while for the latter, we provide details on our handling of RoPE in Appendix~\ref{apd:token_merging_rope}. Here, we present only the ablation studies on the merging ratio on Table~\ref{tab:merging_ratio_comparison}. The comparison experiments are provided in Appendix~\ref{apd:comparison_tomemgt}.

\input{tables/tomemgt_merging_ratio}

\subsection{Momentum-based Solver}
\label{sec:momentum_based_solver}

Inspired by the success of DDIM~\citep{ddim} and DPM-Solver~\citep{dpm_solver} in DM, we aim to implement a similar mechanism in MGT. Since the 1st order form of DPM-Solver is equivalent to DDIM, we focus on implementing DPM-Solver. We call our implementation of DPM-Solver in MGT Momentum-based Solver since these algorithms basically use momentum for accelerated sampling~\citep{dpm_solver_v3,iclr22_progressive,icml23_consistency}. The analytical equations of the 1st and 2nd orders can be written as\begin{equation}
    \begin{split}
        & \ve{z}_t = \frac{\sigma_t}{\sigma_s} \ve{z}_s - \frac{\sigma_t-\sigma_s}{\sigma_s} f_\theta(\ve{z}_s,\sigma_s), \text{\textcolor{C3}{\quad\quad\quad\# 1st order}} \\
        & \ve{z}_t = \frac{\sigma_t}{\sigma_s}\ve{z}_s - \frac{\sigma_t-\sigma_s}{\sigma_s} f_\theta(\ve{z}_s,\sigma_s) + \text{\textcolor{C3}{\quad\quad\# 2nd order}} \\
        &\quad\quad\left((\sigma_t-\sigma_s)^2/2\sigma_t\right)\left(\partial f_\theta(\ve{z}_s,\sigma_s)/\partial \sigma_s\right),
    \end{split}
    \label{eq:momentum_based_solver}
\end{equation}where $t,s \in \{0,\cdots,N\}$ and $1\!\leq\!s\!<\!t\!\leq\!N$. Our derivation can be found in Appendix~\ref{apd:derivation_momentum_based_solver}. The challenge of using Eq.~\ref{eq:momentum_based_solver} lies in performing addition/subtraction operations on the token maps at different time steps. We adopt a simple yet effective approach by transforming the operations into probability distributions and then performing token replacement based on these distributions. For instance, since $\frac{\sigma_t}{\sigma_s}\!+\!\left(-\frac{\sigma_t - \sigma_s}{\sigma_s}\right)\!=\!1$, for the 1st order Solver, we select $100\frac{\sigma_t}{\sigma_s}\%$ of the tokens from $\ve{z}_s$ and $100\frac{\sigma_s - \sigma_t}{\sigma_s}\%$ of the tokens from $f_\theta(\ve{z}_t, \sigma_t)$, and simply merge them.
\begin{figure}[!h]
\vspace{-8pt}
\centering
\hspace{-0.5cm}
\includegraphics[width=0.5\textwidth,trim={0cm 0.0cm 0cm 0cm},clip]{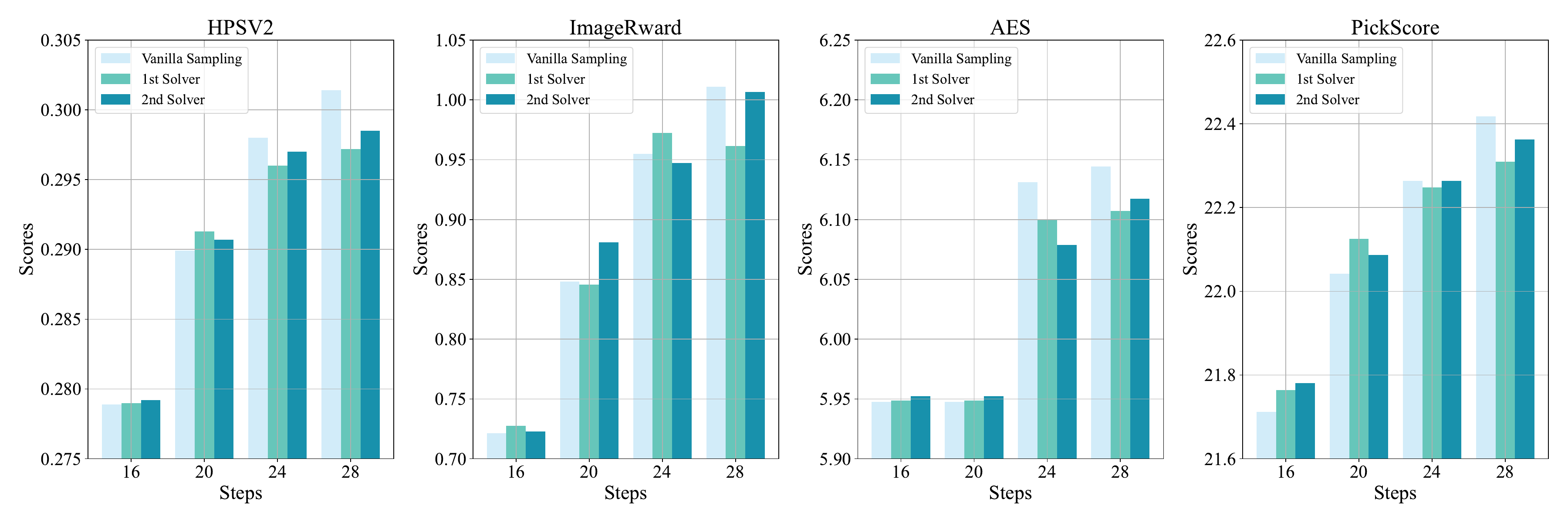}
\vspace{-20pt}
\caption{Ablation Studies of Momentum-based Solver.}
\vspace{-10pt}
\label{figure:momentum_based_solver_ablation}
\end{figure}
For the token selection rule, we follow a high-confidence criterion, selecting the top $100\frac{\sigma_s - \sigma_t}{\sigma_s}\%$ of tokens with the highest confidence from $f_\theta(\ve{z}_t, \sigma_t)$. Additionally, for the gradient in the 2nd order solver, we use the same difference expansion form as in DPM-Solver. We present the ablation experiments in Fig.~\ref{figure:momentum_based_solver_ablation}, which reveals that the Momentum-based Solver provides performance gains for $N$=16 and $N$=20, but does not perform as well as vanilla sampling for larger $N$. We argue this is due to the discrete nature of the token values, which restricts the effectiveness of addition/subtraction operations, unlike in DM.

%% file: tables/karras_inv_sigma_01.tex
\begin{table}[!h]
\centering
\vskip -0.01in
{\small
\renewcommand\arraystretch{0.95}
\setlength{\tabcolsep}{3.8pt}
\vspace{-3pt}
\hspace{-0.4cm}
\resizebox{.48\textwidth}{!}{%
\begin{tabular}{lcccc}
\hline
Method & HPS v2 ($\uparrow$) & ImageReward ($\uparrow$) & AES ($\uparrow$) & PickScore ($\uparrow$) \\ \hline
{\footnotesize Cosine Schedule} & 0.3062 & 1.0627 & 6.1938 & 22.5037 \\
$1-t^\rho$ ($\rho=1$) & 0.2835 & 0.7679 & 6.0017 & 21.8594 \\
$1-t^\rho$ ($\rho=0.95$) & 0.2866 & 0.8160 & 6.0652 & 21.9837 \\
$1-t^\rho$ ($\rho=0.9$) & 0.2914 & 0.8647 & 6.0632 & 22.1004 \\
$1-t^\rho$ ($\rho=0.85$) & 0.2939 & 0.9500 & 6.1004 & 22.2132 \\
$1-t^\rho$ ($\rho=0.8$) & 0.2974 & 0.9728 & 6.1094 & 22.3202 \\
$1-t^\rho$ ($\rho=0.75$) & 0.3011 & 1.0323 & 6.1785 & 22.3804 \\
$1-t^\rho$ ($\rho=0.7$) & 0.3019 & 1.0407 & 6.1734 & 22.4451 \\
$1-t^\rho$ ($\rho=0.65$) & 0.3051 & 1.0153 & 6.1945 & 22.4474 \\
$1-t^\rho$ ($\rho=0.6$) & \CC 0.3080 & \CC 1.0889 & 6.2101 & \CC 22.5380 \\
$1-t^\rho$ ($\rho=0.55$) & 0.3051 & 1.0520 & \CC 6.2351 & 22.4428 \\
$1-t^\rho$ ($\rho=0.5$) & 0.3052 & 1.0476 & 6.2155 & 22.4381 \\
\bottomrule
\end{tabular}}
\caption{\small {Comparison between cosine schedule and $1-t^\rho$.}}
\label{tab:karras_sigma_01}}
\vspace{-8pt}
\end{table}

%% file: tables/model_quantization_ablation_study.tex
\begin{table*}[t]
\centering
\renewcommand\arraystretch{0.8}
\setlength{\tabcolsep}{10pt}
\resizebox{1.0\textwidth}{!}{%
\begin{tabular}{lcccccc}
\hline
{Method} & {Total Memory Usage} & {Transformer Memory Usage} & {HPS v2 (↑)} & {ImageReward (↑)} & {AES (↑)} & {PickScore (↑)} \\
\hline
Float32 & 11.98 GB & 6.82 GB & 0.3062 & 1.0627 & 6.1938 & 22.5037 \\
Float16 & 9.55 GB & 3.38 GB & 0.3041 & 1.0019 & 6.1827 & \CC 22.5333 \\
Bfloat16 & 6.78 GB & 3.34 GB & \CC 0.3069 & \CC 1.0867 & \CC 6.2222 & 22.4525 \\
A8W4-PTQ & 5.86 GB & \CC 2.24 GB & 0.1009 & -2.2533 & 4.1185 & 17.1369 \\
\makecell[l]{A8W4-QAT \& Calibration} & 5.86 GB & \CC 2.24 GB & 0.3063 & 1.0055 & 6.1227 & 22.2648 \\
\makecell[l]{A8W4-SCQ (ours)} & 5.86 GB & \CC 2.24 GB & 0.3066 & 1.0635 & 6.1907 & 22.4406 \\
\makecell[l]{A8W4-SCQ (CPU offloading, ours)} & \CC 4.57 GB & \CC 2.24 GB & 0.3066 & 1.0635 & 6.1907 & 22.4406 \\
\hline
\end{tabular}}
\vspace{-5pt}
\caption{Comparison of different quantization methods on HPD v2 Subset. Our proposed SCQ reduces total memory usage from 11.98 GB to 4.57 GB with minimal performance loss. All memory is recorded using \texttt{torch.cuda.max\_memory\_reserved()}.}
\label{tab:memory_comparison}
\vspace{-8pt}
\end{table*}

%% file: tables/challengebench.tex

\begin{table*}[!t]
\begin{center}
\footnotesize
\renewcommand\arraystretch{0.85}
\setlength{\tabcolsep}{10pt}
\vspace{-0.06in}
\resizebox{1.0\linewidth}{!}{
\begin{tabular}{ccccccc}
\hline
              T2I-Compbench           & \multicolumn{2}{c}{Attribute Binding}                                                                                       & \multicolumn{3}{c}{Object Relationship}                                                                                                                               &                                         \\ \cline{2-6}
(Meissonic) & Color~(↑)   & Texture~(↑)                                 & 2D-Spatial~(↑)                              & 3D-Spatial~(↑)                              & Non-Spatial~(↑)  & \multirow{-2}{*}{Complex~(↑)}               \\ \hline
Vanilla Sampling & 0.5540	& 0.4858 & 0.1809 & 0.3381 & 0.3037 & 0.2942 \\
Noise Regularization & \CC {0.5682} & 0.4937 & \CC {0.1932} & \CC {0.3687} & \CC {0.3053} & 0.2976 \\ 
Differential Sampling & 0.5458 & 0.4561 & 0.1515 & 0.3598 & 0.3041 & \CC {0.2994} \\ 
Masked Z-Sampling & 0.5451 & \CC {0.5106} & 0.1738 & 0.3642& 0.3033 & 0.2913 \\\hline
\end{tabular}
}
\end{center}
\vspace{-16pt}
\caption{Comparison of the combination of our methods on T2I-Compbench~\citep{t2i_compbench} with Meissonic-1024$\times$1024.}
\vspace{-7pt}
\label{figure:t2i_compbench}
\end{table*}\begin{table*}[!t]
\begin{center}
\footnotesize
\renewcommand\arraystretch{0.8}
\setlength{\tabcolsep}{10pt}
\vspace{-0.1in}
\resizebox{1.0\linewidth}{!}{\begin{tabular}{cccccccc}
\hline
              GenEval         & \multirow{2}{*}{Single Object~(↑)} & \multirow{2}{*}{Two Object~(↑)} & \multirow{2}{*}{Counting~(↑)} & \multirow{2}{*}{Colors~(↑)} & \multirow{2}{*}{Position~(↑)} & \multirow{2}{*}{Color Attr~(↑)} & \multirow{2}{*}{Avg.~(↑)}     \\
(Meissonic) &  & & & & & &  \\ \hline
Vanilla Sampling & 91.25\% & 54.61\% & 37.50\% & 79.91\% & 9.00\% & 15.00\% & 47.87\% \\
Noise Regularization & \CC {95.00\%} & 51.52\% & \CC {41.25\%} & 80.85\% & 5.00\% & 20.00\% & 48.93\% \\
Differential Sampling & 92.50\% & 54.44\% & 33.75\% & 79.66\% & \CC {13.00\%} & 15.00\% & 48.05\% \\
Masked-Z Sampling & 93.75\% & \CC {55.56\%} & 36.25\% & \CC {84.04\%} & 9.00\% & \CC {26.00\%} & \CC {50.77\%} \\ \hline
\end{tabular}
}
\end{center}
\vspace{-16pt}
\caption{Comparison of the combination of our methods on GenEval~\citep{geneval} with Meissonic-1024$\times$1024.}
\vspace{-7pt}
\label{figure:geneval}
\end{table*}\begin{figure*}[!t]
\vspace{-3pt}
\centering
\hspace{-0.5cm}
\includegraphics[width=0.85\textwidth,trim={0cm 0.0cm 0.5cm 1cm},clip]{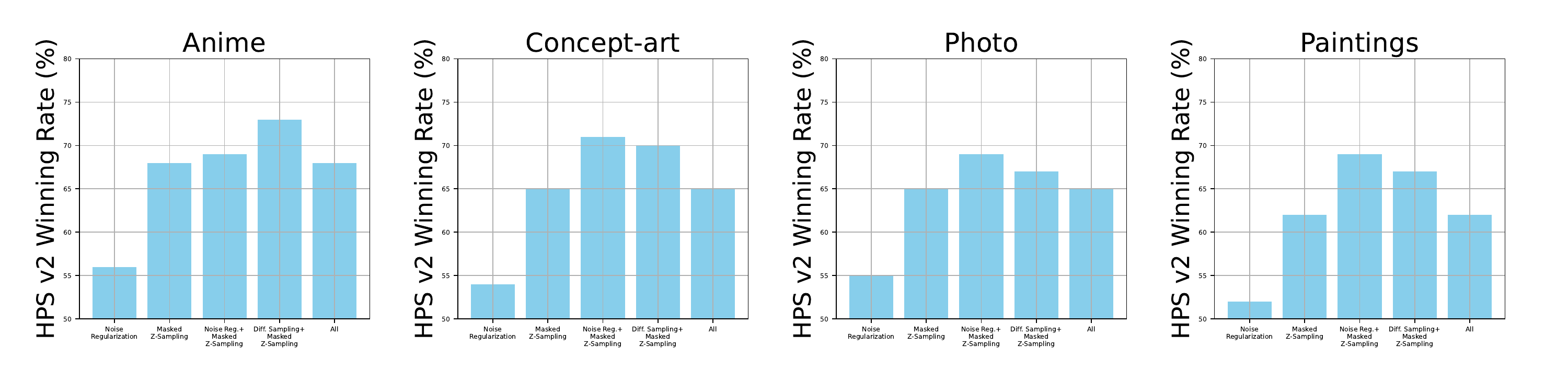}
\vspace{-16pt}
\caption{Comparison of the combination of our methods on HPD v2. The results for Meissonic-512$\times$512 can be found in Appendix~\ref{apd:experiments_meissonic_512}.}
\vspace{-10pt}
\label{figure:combination_hpsv2}
\end{figure*}

%% file: tables/tomemgt_merging_ratio.tex
\begin{table}[t]
\centering
\renewcommand\arraystretch{0.8}
\vspace{-3pt}
\resizebox{0.48\textwidth}{!}{%
\begin{tabular}{lccccc}
\hline
{Merging Ratio} & {HPS v2} & {Image Reward} & {AES} & {PickScore} & \makecell{Time Spent\\(s/per img)} \\
\hline
0\% & \CC 0.3062 & \CC 1.0627 & \CC 6.1938 & \CC 22.5037 & 8.38 \\
25\% & 0.2976 & 0.9187 & 6.0479 & 22.1302 & 7.98 \\
50\% & 0.2723 & 0.6365 & 5.8335 & 21.3290 & 7.97 \\
75\% & 0.1182 & -2.1480 & 4.1857 & 17.6813 & \CC 7.90 \\
\hline
\end{tabular}}
\vspace{-8pt}
\caption{Ablation studies of merging ratio on a single RTX 4090.}
\label{tab:merging_ratio_comparison}
\vspace{-15pt}
\end{table}

%% file: sec/experiment.tex
\section{Additional Evaluations}
\label{sec:challengebench}

\input{tables/maskgit_main_paper}
To substantiate the generalization ability of the enhanced inference algorithm proposed by us, we conduct experiments on more MGTs (\textit{i.e.}, Meissonic-512$\times$512 and MaskGIT-512$\times$512) as well as more benchmarks (\textit{i.e.}, HPD v2, GenEval and T2I-Compbench). The experimental results on Meissonic-1024$\times$1024 within T2I-Compbench~\citep{t2i_compbench}, GenEval~\citep{geneval}, and HPD v2 are summarized in Table~\ref{figure:t2i_compbench}, Table~\ref{figure:geneval}, and Fig.~\ref{figure:combination_hpsv2}, respectively. These results clearly highlight the effectiveness of all proposed design strategies. Specifically, when integrated with the other two strategies, masked Z-Sampling achieves winning rates of approximately 70\% compared to vanilla sampling on HPD v2. For clarity and due to space limitations in the main paper, the remain experimental results for Meissonic-512$\times$512 and Meissonic-1024$\times$1024 are provided in Appendix~\ref{apd:experiments_meissonic_512}. We further validate the significant performance of our proposed design strategies on class-to-image (C2I) MaskGIT~\citep{maskgit}. As presented in Table~\ref{tab:maskgit_main_paper}, all strategies contribute to performance gains, with masked Z-Sampling yielding the most notable improvements. Moreover, we demonstrate that these strategies are capable of refining token distributions even on ARM (\textit{e.g.}, LlamaGen~\citep{llamagen}). For further details, please refer to Appendix~\ref{apd:llamagen}.


%% file: tables/maskgit_main_paper.tex
\begin{table}[!h]
\vspace{-8pt}
\footnotesize
\renewcommand\arraystretch{0.95}
\setlength{\tabcolsep}{10pt}
\vspace{-0.06in}
\resizebox{1.0\linewidth}{!}{
\begin{tabular}{cccccc}
\hline
MaskGIT-512$\times$512 & IS~(↑) & FID~(↓) & Prec.~(↑) & Recall~(↑) & sFID~(↓) \\ 
\hline
Vanilla Sampling & 291.30 & 14.55 & 0.8403 & 0.165 & 39.26 \\
Noise Regularization & 294.72 & 14.18 & 0.8666 & 0.184 & 36.90 \\
Differential Sampling & 291.30 & 14.35 & 0.8658 & 0.166 & 39.27 \\
Masked-Z Sampling & \CC {298.12} & \CC {12.87} & \CC {0.8842} & \CC {0.199} & \CC {33.58} \\
\hline
\end{tabular}
}
\vspace{-9pt}
\caption{Comparison results of MaskGIT on traditional metrics.}
\label{tab:maskgit_main_paper}
\vspace{-15pt}
\end{table}

%% file: sec/conclusion.tex
\section{Conclusion}
\label{sec:conclusion}

Our approach with the masked generative Transformer, aimed at ensuring enhanced and efficient inference, represents a meaningful exploration of non-autoregressive models. In future, we will try to unify and improve the training process of the masked generative Transformer and overcome the bottlenecks of this generative paradigm. 

%% file: sec/suppl.tex
\clearpage
\setcounter{page}{1}
\maketitlesupplementary

\section{Benchmark and Evaluation Metrics}
\label{apd:benchmark}

In this section, we provide an overview of the benchmarks, evaluation metrics, and related content used in our main paper.

\subsection{Benchmark}

\paragraph{HPD v2.} The Human Preference Dataset v2~\citep{HPSV2} is a large-scale dataset with clean annotations that focuses on user preferences for images synthesized from text prompts. It consists of 798,090 binary preference choices across 433,760 image pairs, designed to address the shortcomings of existing evaluation metrics that do not accurately capture human preferences. In line with the methodologies outlined in~\citep{HPSV2,meissonic}, we utilized four subsets for our review: Animation, Concept-art, Painting, and Photo, with each subset containing 800 prompts.

\paragraph{HPD v2 Subset.} In order to reduce the evaluation computational overhead, we randomly selected 150 prompts from HPD V2~\citep{HPSV2} to build a new collection of prompts for evaluation.

\paragraph{Challengebench.} This benchmark was proposed by us, with the detailed prompt collection methodology described in Appendix~\ref{apd:challengebench_apd}. The core objective of this benchmark is to select challenging prompts, thereby enabling an exploration of the performance limits of generative models. We initially generated 150,000 images using SD XL Base v1.0. We then filtered out the lowest-performing 1,500 prompts based on HPS v2 metrics. Finally, through a combination of manual filtering and GPT-4o filtering, we arrived at a final set of 220 semantically correct and complete prompts.

\paragraph{GenEval.} GenEval is an object-focused framework designed to evaluate compositional image properties, including object co-occurrence, position, count, and color. This benchmark utilizes state-of-the-art object detection models to assess text-to-image generation tasks, ensuring strong alignment with human agreement. Additionally, other discriminative vision models can be integrated into this pipeline to further verify attributes such as object color. Notably, this benchmark comprises 550 prompts, each of which is straightforward and easy to understand.

\paragraph{T2I-Compbench.} T2I-Compbench is a benchmark similar to GenEval, designed as a comprehensive evaluation framework for open-world compositional text-to-image synthesis. It comprises 6,000 compositional text prompts, categorized into three main groups (attribute binding, object relationships, and complex compositions) and further subdivided into six subcategories: color binding, shape binding, texture binding, spatial relationships, non-spatial relationships, and complex compositions.

\subsection{Evaluation Metrics}

\paragraph{PickScore.} PickScore is a scoring function based on the CLIP model, developed using the Pick-a-Pic dataset, which gathers user preferences for generated images. This method demonstrates performance that exceeds typical human benchmarks in predicting user preferences. By effectively aligning with human evaluations and utilizing the diverse distribution prompts from Pick-a-Pic, PickScore facilitates a more pertinent assessment of text-to-image models compared to conventional metrics like FID~\citep{fid} on datasets such as MS-COCO~\citep{COCO}.

\paragraph{HPS v2.} The Human preference score version 2 (HPS v2) represents a refined model for predicting user preferences, achieved through fine-tuning the CLIP framework~\citep{CLIP} on the human preference dataset version 2. This model is designed to align text-to-image generation with user tastes by estimating the likelihood that a generated image will be favored by individuals, thereby serving as a robust tool for evaluating the efficacy of text-to-image models across varied image distributions.

\paragraph{AES.} The Aesthetic score (AES)~\citep{AES} is computed using a model built on CLIP embeddings, enhanced with additional multilayer perceptron (MLP) layers to quantify the visual appeal of images. This metric is useful for assessing the aesthetic quality of generated images, offering insights into their alignment with human aesthetic preferences.

\paragraph{ImageReward.} ImageReward~\citep{Imagereward} is a specialized reward model focused on evaluating text-to-image synthesis through human preferences. It is trained on a substantial dataset comprising human comparisons, enabling it to effectively capture user inclinations. The model evaluates synthesized images based on several factors, including their correspondence to the text prompt and overall aesthetic merit. ImageReward has demonstrated superior performance over traditional metrics such as the Inception Score (IS)~\citep{is} and Fréchet Inception Distance (FID) in reflecting human judgments, positioning it as a promising automatic evaluation tool for text-to-image synthesis.

\paragraph{CLIPScore.} CLIPScore~\citep{CLIPScore} utilizes the strengths of the CLIP model, which integrates images and text within a shared embedding space. By measuring the cosine similarity between image and text embeddings, CLIPScore provides an evaluation of how closely a generated image aligns with its textual description. Although effective in assessing text-image correlation, CLIPScore may not fully capture the subtleties of human preferences, especially regarding aesthetic qualities and intricate details.

\section{Ineffective Method}
\label{apd:ineffective_method}

\begin{figure*}[t]
\vspace{-8pt}
\centering
\includegraphics[width=0.95\textwidth,trim={0cm 0cm 0cm 0cm},clip]{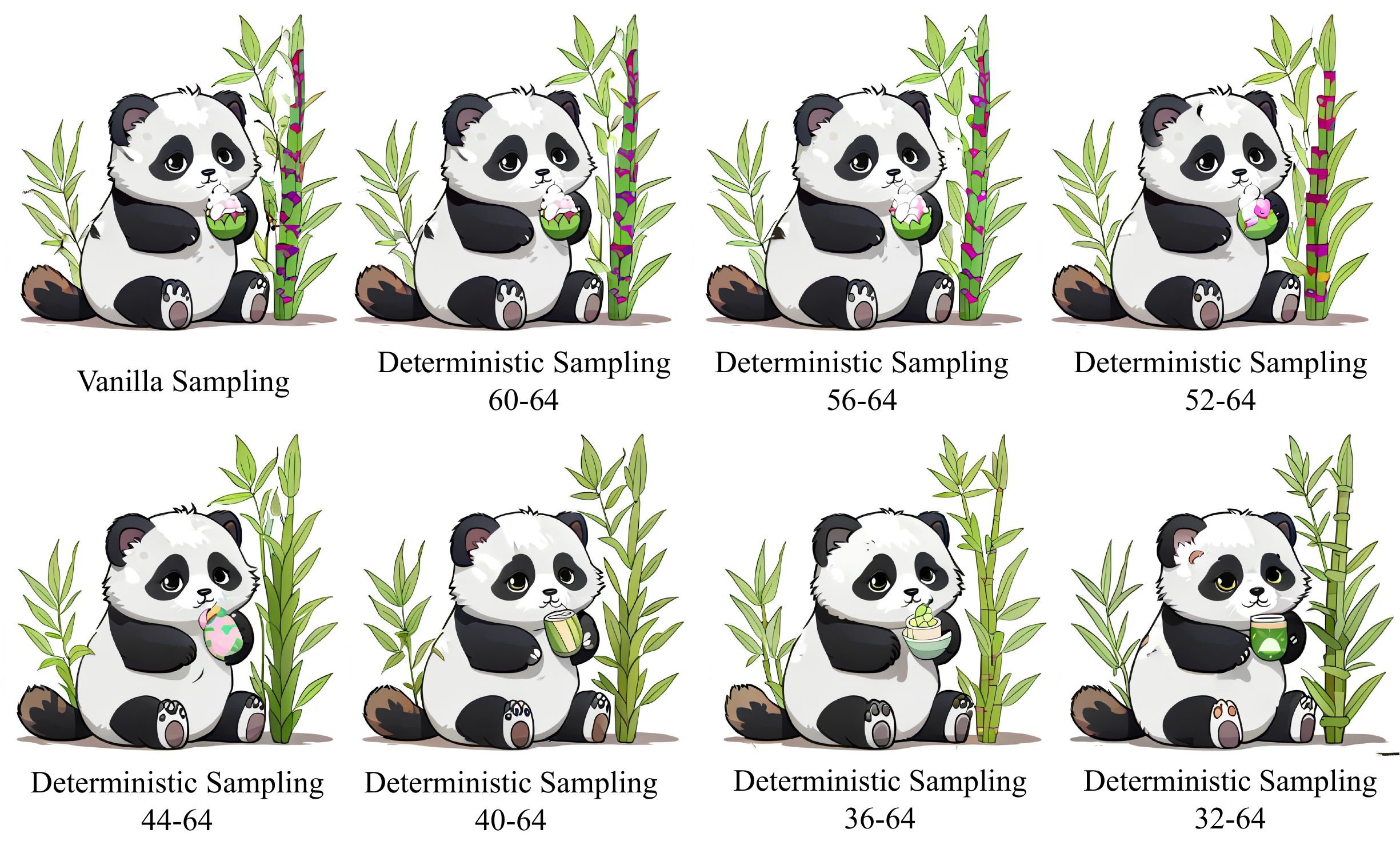}
\vspace{-6pt}
\caption{The visualization of deterministic sampling, where the number following \texttt{deterministic sampling} indicates the step from which \texttt{torch.argmax} is used in place of \texttt{torch.multinomial}, results in outputs that do not exhibit significant perceptual quality degradation. Despite the lack of noticeable decline in visual fidelity, this shift adversely affects the quantitative metrics.}
\vspace{-12pt}
\label{figure:deterministic_sampling}
\end{figure*}

Here, we summarize a collection of algorithms that demonstrate limited effectiveness when applied to MGT, aiming to provide valuable insights for other researchers.

\subsection{Deterministic Sampling}
\label{apd:deterministic_sampling}

Deterministic sampling techniques, such as DDIM~\citep{ddim}, have been developed for DMs to facilitate tasks like image editing and accelerated sampling. Consequently, we are interested in exploring whether MGT can be adapted for deterministic sampling. In our investigation, we identified two stochastic elements within the sampling mechanism of MGT: first, the process of determining which regions of the subsequent sample should be masked; second, the sampling procedure defined by $\arg\max_i \frac{\log(\epsilon)}{\ve{p}}$, where $\ve{p}$ represents the logit and $\epsilon$ is drawn from a uniform distribution $\mathcal{U}[\ve{0}, \ve{1}]$. We find that eliminating randomness in the former causes the sampling pipeline to collapse. In the latter, randomness can only be reduced during the later stages of sampling (\textit{i.e.}, when $t$ approaches 1); otherwise, the sampling quality deteriorates significantly.

\input{tables/deterministic_sampling}

Table~\ref{tab:deterministic_sampling} presents the results, showing that the introduction of deterministic sampling does not lead to significant metric improvements but instead results in performance degradation. Consequently, we do not consider it a valid technique.

\section{Additional Information of Effective Method}
\label{apd:additional_information}

Here, we present discussions, analyses, and experimental results that could not be developed due to space limitations in the main paper.

\subsection{Additional Experiments of Different Noise Schedule}
\label{sec:additional_experiment_noise_schedule}
\begin{figure}[!h]
\vspace{-5pt}
\centering
\hspace{-0.5cm}
\includegraphics[width=0.5\textwidth,trim={0cm 0.0cm 0cm 0cm},clip]{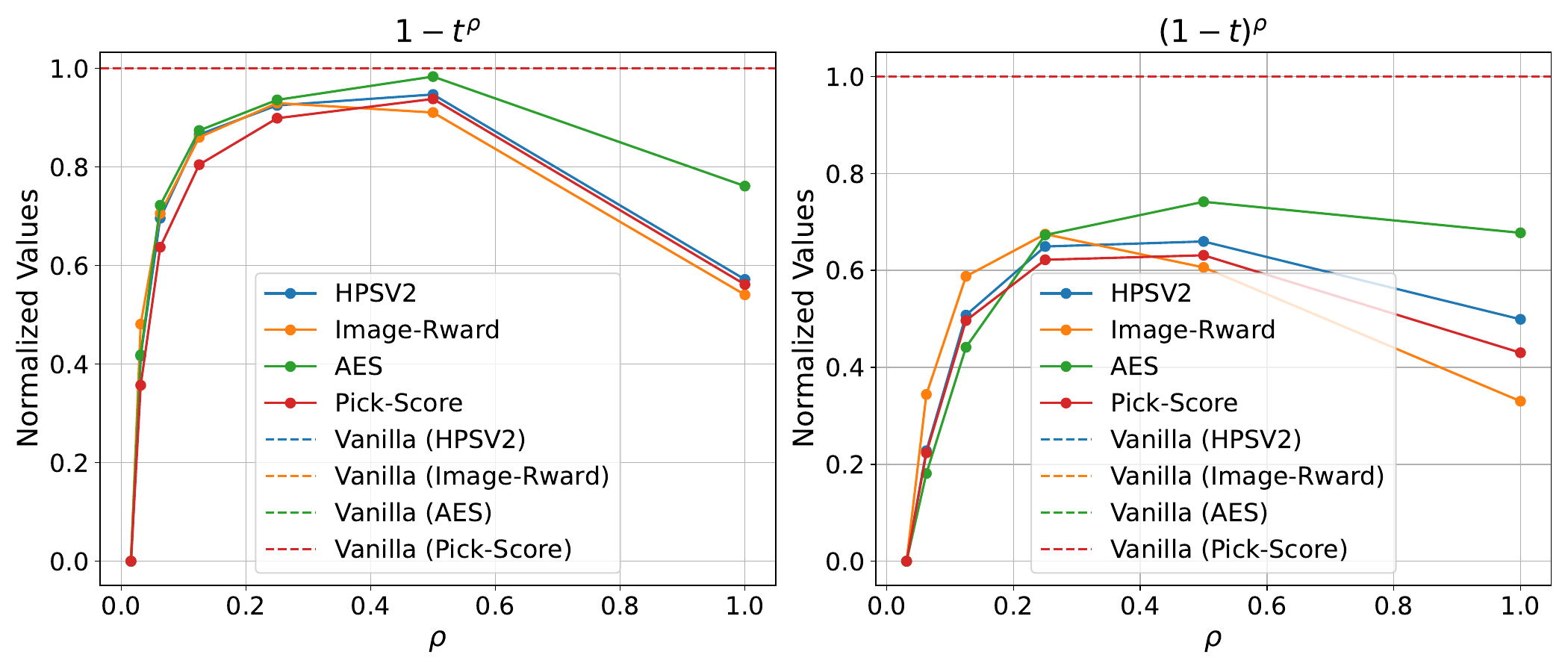}
\vspace{-10pt}
\caption{Visualization of the performance of different noise schedules. The dotted line denotes the vanilla sampling, and $N$ is set to 48.}
\vspace{-5pt}
\label{figure:karras_sigma_result_n48}
\end{figure}\begin{figure}[!h]
\vspace{-5pt}
\centering
\hspace{-0.5cm}
\includegraphics[width=0.5\textwidth,trim={0cm 0.0cm 0cm 0cm},clip]{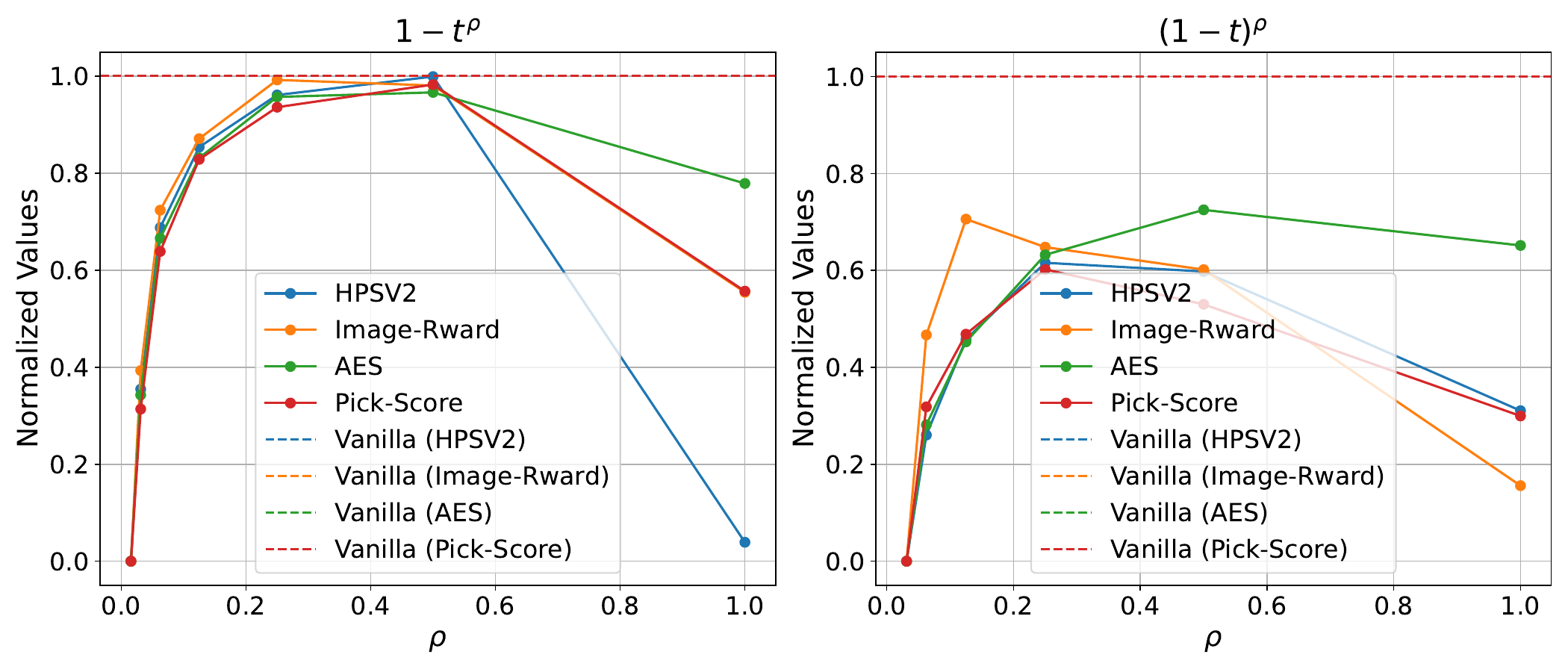}
\vspace{-10pt}
\caption{Visualization of the performance of different noise schedules. The dotted line denotes the vanilla sampling, and $N$ is set to 32.}
\vspace{-5pt}
\label{figure:karras_sigma_result_n32}
\end{figure}\input{tables/karras_inv_sigma_01_full_HPD}
As a supplement to Sec.~\ref{sec:convex_exploration}, we present the experimental results for the number of sampling steps $N = 48$ and $N = 32$, conducted on the HPD v2 Subset, in Figs~\ref{figure:karras_sigma_result_n48} and~\ref{figure:karras_sigma_result_n32}, respectively. We further tested the performance of the noise schedule $1-t^{0.6}$  at $N = 64$ on the full HPD v2 benchmark and presented the result in Table~\ref{tab:inv_karras_comparison_fullHPD}. Table~\ref{tab:inv_karras_comparison_fullHPD} shows that $1-t^{0.6}$ outperforms $\mathrm{cos}(\frac{\pi t}{2})$ on the vast majority of metrics and subsets within the HPD v2, despite the potential risk of overfitting introduced by this exhaustive search approach on the hyperparameter $\rho$.

\subsection{Analysis of Noise Regularization and Differential Sampling}
\label{apd:analysis_noise_reg}
\begin{figure}[h]
\vspace{-8pt}
\centering
\hspace{-0.5cm}
\includegraphics[width=0.5\textwidth,trim={0cm 0.0cm 0cm 0cm},clip]{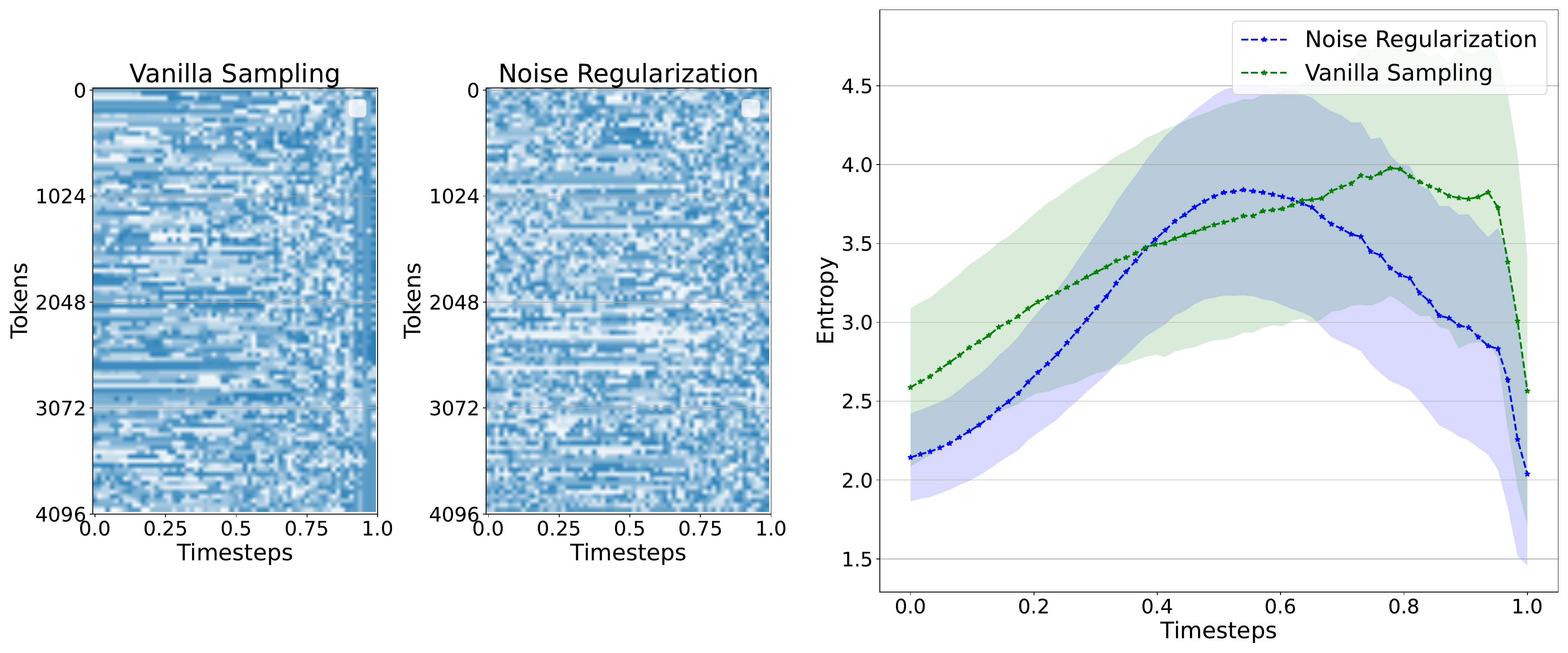}
\vspace{-20pt}
\caption{\textbf{Left:} The sampling trajectory of noise regularization. Different colored tokens represent different values $\in \{0,\cdots,8191\}$. \textbf{Right:} The visualization of the entropy as the sampling progresses. Shaded blocks are the standard deviations.}
\vspace{-8pt}
\label{figure:noise_regularization_entropy_traj}
\end{figure}
Here, we present empirical arguments for why noise regularization works. As shown in Fig.~\ref{figure:noise_regularization_entropy_traj} (Left), vanilla sampling often results in horizontal bars of the same color as $t$ approaches 0 and vertical bars of the same color as $t$ approaches 1, indicating the redundancy of its sampling process. To be specific, we calculate the mean values of the absolute differences between adjacent horizontal bars and vertical bars: standard sampling is $\frac{43.2}{255}$ and $\frac{27.0}{255}$, while noise regularization is $\frac{45.6}{255}$ and $\frac{35.3}{255}$. The increased variability in bar intensities under noise regularization—evidenced by higher mean differences—quantitatively confirms the method’s efficacy in reducing same-color bar artifacts. Interestily, as presented in Fig.~\ref{figure:noise_regularization_entropy_traj} (Right), noise regularization modifies the entropy of the distribution during the sampling phase, making it more inclined toward a ``U''-shaped structure, which is similar to the shape of $|\cos(\pi t)|$.

\subsection{Exploration of Z-Sampling}
\label{apd:exploration_zigzag}
In this subsection we focus on showing some experimental results of Z-Sampling as a complement to the main paper. 
\input{tables/zigzag_sampling_apd_1}
The results in Table~\ref{tab:random_masking_mechanism} are obtained using the random masking mechanism (\textit{i.e.}, vanilla Z-Sampling in Fig.~\ref{figure:zigzag_sampling_illustration}), with the numbers in parentheses indicating the total number of Z-Sampling operations performed from the 0th step. It can be observed that this approach generally impairs the inference performance of the MGT.

\input{tables/zigzag_sampling_apd_2}

We also designed another masking algorithm named recover Z-sampling, which directly reuses the mask from the first sampling from the $(i\!-\!1)$th step to the $i$th step. The results of vanilla sampling, this algorithm and masked Z-Sampling are presented in Table~\ref{tab:zigzag_sampling_recover_improved}, where we find that this approach also yields significant performance gains, similar to masked Z-Sampling. It should be noted that, unless otherwise specified, all experiments in this paper are conducted using masked Z-Sampling.

\subsection{Comparison of Noise Regularization}
\label{apd:comparison_noise_regularization}

\input{tables/noise_regularization_apd_1}
We present the results of the comparative experiments on noise regularization in Table~\ref{tab:noise_regularization_comparison}, which show that it outperforms vanilla sampling across all domains and metrics.

\subsection{Comparison of Differential Sampling}
\label{apd:comparison_differential_sampling}

\input{tables/differential_sampling_apd_1}
Similar to Sec.~\ref{apd:comparison_noise_regularization}, we present the results of the comparative experiments on differential sampling in Table~\ref{tab:differential_sampling_comparison}. On HPD v2, a large-scale benchmark consisting of 3,200 prompts, differential sampling outperforms vanilla sampling across nearly all metrics and domains.

\subsection{Detail Implementation of Token Merging on RoPE}
\label{apd:token_merging_rope}

Here, we describe how to perform token merging on RoPE. Assume that tokens can be defined as $\{\ve{n}_i\}_{i=1}^N$, then the self-attention can be expressed as
\begin{equation}
    \begin{split}
        & q_i = f_q(\ve{n}_i, i),\\
        & k_j = f_k(\ve{n}_j, j),\\
        & v_j = f_v(\ve{n}_j, j), \\
        & a_{i,j} = \frac{\mathrm{exp}(\frac{q^T_ik_j}{\sqrt{d}})}{\sum_{k=1}^N\mathrm{exp}(\frac{q^T_ik_k}{\sqrt{d}})} \\
        & \ve{o}_j = \sum_{j=1}^N a_{i,j}v_j,
    \end{split}
\end{equation}
where $N$ and $d$ denote the number of tokens and the length of each token, respectively. The core of RoPE is to inject positional information into the computations involving $f_q(\cdot,\cdot)$ and $f_k(\cdot,\cdot)$, the key lies in the matrix $Q^d_j$:
\begin{equation}
\tiny
\begin{split}
& Q_j^d = \\
&\left(
  \begin{array}{ccccc}
    \mathrm{cos}(j\theta_0) & -\mathrm{sin}(j\theta_0) & \cdots & 0 & 0 \\
    \mathrm{sin}(j\theta_0) & \mathrm{cos}(j\theta_0) & \cdots & 0 & 0 \\
    0 & 0  & \cdots & 0 & 0 \\
    0 & 0  & \cdots & 0 & 0 \\
    \vdots & \vdots & \ddots & \vdots & \vdots \\
    0 & 0 & \cdots & \mathrm{cos}(j\theta_{\frac{d}{2}-1}) & -\mathrm{sin}(j\theta_{\frac{d}{2}-1}) \\
    0 & 0 & \cdots & \mathrm{sin}(j\theta_{\frac{d}{2}-1}) & \mathrm{cos}(j\theta_{\frac{d}{2}-1}) \\
  \end{array}
\right),\\
\end{split}
\end{equation}
where $\theta_i = 10000^{-2(i-1)/d}$ and $i\in [1,2,\cdots, d/2]$. The functions $f_q(\cdot, \cdot)$ and $f_k(\cdot, \cdot)$ can then be expressed as
\begin{equation}
    \begin{split}
        & f_q(n_j, j) = Q_j^dW_q \ve{n}_j,\\
        & f_k(n_j, j) = Q_j^dW_k \ve{n}_j.\\
    \end{split}
\end{equation}
TomeMGT first computes a similarity matrix between tokens to determine which highly similar tokens should be merged. For the tokens identified for merging, their mean value is calculated to obtain the merged tokens. Given the tokens $\ve{n}_i$ and $\ve{n}_j$ that need to be merged, their merged token can be obtained as $\ve{n}_{i+j} = \frac{\ve{n}_i + \ve{n}_j}{2}$. We found it effective to apply the same strategy for RoPE, specifically, $Q_{i+j}^d = \frac{Q_i^d + Q_j^d}{2}$. It is worth noting that we also tried using $Q_{i+j}^d$ directly as either $Q_{i}^d$ or $Q_{j}^d$, but this approach led to inference collapse.

\subsection{Comparison of TomeMGT}
\label{apd:comparison_tomemgt}

\input{tables/tomemgt_apd_1}
Here, we provide the experimental results of our proposed TomeMGT on HPD v2, as shown in Table~\ref{tab:tomemgt_comparison}. The merging ratio is set to 0.5 for all experiments. It can be observed that TomeMGT does not perform as well compared to our proposed methods: noise regularization, differential sampling, masked Z-Sampling, and SCQ. However, this algorithm may hold significant potential if the parameter size and number of tokens in MGT are further increased in the future.

\subsection{Derivation of Momentum-based Solver}
\label{apd:derivation_momentum_based_solver}

The original derivation of DPM-Solver was based on the DM paradigm, not MGT. However, DPM-Solver is theoretically adaptable to DMs with any noise schedule, which suggests the potential for its application to MGT. While MGT uses a cosine schedule as the default noise schedule, other forms, such as the Karras-like schedule introduced in this paper~\citep{karras2022elucidating}, can also be applied. The crucial requirement in MGT is that $\alpha_t + \sigma_t = 1$, since the number of masked tokens and the number of unmasked tokens must together equal the total number of tokens. Given this, we first derive the DPM-Solver for predicting $\ve{z}_N$ scenarios based on the flow matching paradigm~\citep{iclr22_rect,icml23_curvature}, and then substitute the corresponding noise schedule. The flow matching can be written as
\begin{equation}
    \begin{split}
        & \frac{d\ve{z}_t}{dt} = -\ve{v}_\theta(\ve{z}_t,t). \\
    \end{split}
\end{equation}
For simplicity, we assume $t$ is continuous and $t \in [0,1]$, where $\ve{z}_1$ corresponds to the original $\ve{z}_N$ and $\ve{z}_0$ to the original $\ve{z}_0$. Let $\ve{v}_\theta(\ve{z}_t, t)$ denote the model for estimating $\ve{z}_0 - \ve{z}_N$. Substituting in the prediction target $\ve{z}_N$, we can obtain
\begin{equation}
    \begin{split}
        & \frac{d\ve{z}_t}{dt} = -\frac{\ve{z}_t - f_\theta(\ve{z}_t,t)}{t}, \\
        & \implies \frac{d\ve{z}_t}{dt} = \frac{-1}{t}\ve{z}_t + \frac{1}{t} f_\theta(\ve{z}_t,t),\\
    \end{split}
\end{equation}
where $f_\theta(\ve{z}_t,t)$ stands for the backbone (\textit{i.e.}, Transformer) of MGT. Calculate the analytical solution of a partial differential equation over a specified interval $[s,t]$ ($0\leq s < t\leq 1$):
\begin{equation}
    \begin{split}
        & \ve{z}_t = e^{\int_s^t \frac{-1}{\tau} d\tau}\ve{z}_s + \int_s^t\left(e^{\int_\tau^t  \frac{-1}{r} dr}(\frac{1}{\tau}f_\theta(\ve{z}_\tau,\tau)) \right)d\tau, \\
        & \implies \ve{z}_t =\left(\frac{s}{t}\right) \ve{z}_s + \int_s^t\left(\left[\frac{1}{t}\right]f_\theta(\ve{z}_\tau,\tau)  \right)d\tau. \\
    \end{split}
\end{equation}
Let $f_\theta(\ve{z}_\tau,\tau)$ be Taylor-expanded at $s$, we can obtain:
\begin{equation}
    \begin{split}
        & \ve{z}_t =\frac{s}{t} \ve{z}_s + \int_s^t\left(\left[\frac{1}{t}\right]\left[f_\theta(\ve{z}_s,s)+(\tau-s)\frac{\partial f_\theta(\ve{z}_s,s)}{\partial s}\right]  \right)d\tau. \\
    \end{split}
\end{equation}
In this paper, we consider only second-order Taylor expansions. From this, we derive the first-order and second-order expressions for the Momentum-based Solver:
\begin{equation}
    \begin{split}
        & \ve{z}_t = \frac{s}{t}  \ve{z}_s + \frac{t-s}{t}f_\theta(\ve{z}_s,s),\textcolor{C3}{\text{ \# 1st order }} \\
        & \ve{z}_t = \frac{s}{t}  \ve{z}_s + \frac{t-s}{t}f_\theta(\ve{z}_s,s)\textcolor{C3}{\text{ \# 2nd order }} \\
        & + \frac{(t-s)^2}{2t}\frac{\partial f_\theta(\ve{z}_s,s)}{\partial s}. \\
    \end{split}
    \label{eq:dpm_deri}
\end{equation}
Substitute $\sigma_s$ and $\sigma_t$ into Eq.~\ref{eq:dpm_deri}:
\begin{equation}
    \begin{split}
        & \ve{z}_t = \frac{\sigma_s}{\sigma_t}  \ve{z}_s + \frac{\sigma_t-\sigma_s}{\sigma_t}f_\theta(\ve{z}_s,\sigma_s),\textcolor{C3}{\text{ \# 1st order }} \\
        & \ve{z}_t = \frac{\sigma_s}{\sigma_t}  \ve{z}_s + \frac{\sigma_t-\sigma_s}{\sigma_t}f_\theta(\ve{z}_s,\sigma_s)\textcolor{C3}{\text{ \# 2nd order }} \\
        & + \frac{(\sigma_t-\sigma_s)^2}{2\sigma_t}\frac{\partial f_\theta(\ve{z}_s,\sigma_s)}{\partial \sigma_s}. \\
    \end{split}
\end{equation}

\input{tables/meissonic_512_512_combination}
\input{tables/meissonic_1024_1024_combination}

\input{tables/challengebench_appendix}

\subsection{Verification of Combinations of Algorithms}
\label{apd:experiments_meissonic_512}

Due to space limitations in the main paper, we provide here a combination of design choices from Meissonic-512$\times$512 and Meissonic-1024$\times$1024 for the experiments. As illustrated in Table~\ref{tab:meissonic_512_evaluation} and Table~\ref{tab:meissonic_1024_evaluation} (values in parentheses represent the winning rate of the combined design choice compared to vanilla sampling), it can be observed that these design choices, when combined, produce a synergistic effect greater than the sum of their individual contributions.

\subsection{ChallengeBench}
\label{apd:challengebench_apd}

We further analyze the performance of MGT on challenging prompts. We synthesize 150k images using SD XL~\citep{SDXL}, computed HPS v2 scores, and selected 1.5k prompts with the lowest scores. After manual and GPT-4o~\citep{GPT4} selection, we form \textit{Challengebench} with 220 semantically sound prompts. We conduct experiments using Meissonic, SD XL~\citep{SDXL}, FLUX.1-schnell~\citep{FLUX}, and SD-3.5-Large~\citep{SD35} on Challengebench, and present the results in Table~\ref{tab:challengebench_apd}. We observe that Meissonic's improvement over SD XL on Challengebench (0.1838 $\rightarrow$ 0.2116) was greater than its improvement on HPD v2 (0.2888 $\rightarrow$ 0.2957) in Meissonic's original paper, suggesting that MGT is more robust on challenging prompts. Furthermore, our inference-enhancing algorithms including noise regularization, masked Z-Sampling, and differential sampling continue to have significant performance gains against Meissonic on this benchmark.

\subsection{Additional Experiments on LlamaGen}
\label{apd:llamagen}

Our findings reveal that differential sampling and noise regularization are independent of the inversion operator (masked Z-Sampling cannot implement since lack the inversion operator) and can be directly implemented on ARM. We conducted experiments using LlamaGen~\citep{llamagen}, with the corresponding results shown in Table~\ref{tab:llamagen_1} and Table~\ref{tab:llamagen_2}. Our experiments reveal that these two strategies can significantly improve LlamaGen's generative performance while incurring almost no additional computational overhead. This finding highlights a high alignment in predictive objectives between MGT and ARM.

\input{tables/llamagen}

%% file: tables/deterministic_sampling.tex
\begin{table}[!t]
\centering
\vskip -0.01in
\small
\renewcommand\arraystretch{1.2}
\setlength{\tabcolsep}{10pt}
\scalebox{0.62}{
\begin{tabular}{ccccc}
\hline
Method & HPS v2 (↑) & ImageReward (↑) & AES (↑) & PickScore (↑) \\ \hline
Vanilla & 0.3062 & 1.0627 & 6.1938 & \CC 22.5037 \\
Deter. Sampling (60-64) & \CC 0.3061 & 1.0619 & 6.1983 & 22.4872 \\
Deter. Sampling (56-64) & 0.3056 & 1.0664 & 6.1925 & 22.4618 \\
Deter. Sampling (52-64) & 0.3057 & 1.0498 & 6.1832 & 22.4627 \\
Deter. Sampling (48-64) & 0.3055 & \CC 1.0679 & 6.1823 & 22.4574 \\
Deter. Sampling (44-64) & 0.3048 & 1.0545 & 6.1926 & 22.4675 \\
Deter. Sampling (40-64) & 0.3047 & 1.0676 & 6.1941 & 22.4355 \\
Deter. Sampling (36-64) & 0.3051 & 1.0654 & \CC 6.2043 & 22.4516 \\
Deter. Sampling (30-64) & 0.3043 & 1.0514 & 6.1864 & 22.4204 \\
\bottomrule
\end{tabular}
}
\caption{\small {Quantitative comparison between vanilla sampling and deterministic sampling (Deter. Sampling).}}
\label{tab:deterministic_sampling}
\vskip -0.01in
\end{table}

%% file: tables/karras_inv_sigma_01_full_HPD.tex
\begin{table}[!h]
\centering
\vskip -0.01in
\small
\renewcommand\arraystretch{1.2}
\setlength{\tabcolsep}{10pt}
\resizebox{0.48\textwidth}{!}{%
\hspace{-0.5cm}
\begin{tabular}{lccccc}
\hline
Subset of HPD v2 & Method & HPS v2 ($\uparrow$) & ImageReward ($\uparrow$) & AES ($\uparrow$) & PickScore ($\uparrow$) \\ \hline
Anime & $\mathrm{cos}(\frac{\pi t}{2})$ & 0.3053 & 0.9577 & 6.1049 & 22.4776 \\
Anime & $1-t^{0.6}$ & \CC 0.3099 & \CC 1.0697 & \CC 6.1068 & \CC 22.6601 \\
Photo & $\mathrm{cos}(\frac{\pi t}{2})$ & 0.2658 & 0.5161 & \CC 5.9992 & 21.5552 \\
Photo & $1-t^{0.6}$ & \CC 0.2712 & \CC 0.6011 & 5.9894 & \CC 21.6501 \\
Paintings & $\mathrm{cos}(\frac{\pi t}{2})$ & \CC 0.2915 & 1.0802 & \CC 6.4790 & 21.8669 \\
Paintings & $1-t^{0.6}$ & 0.2907 & \CC 1.0911 & 6.4501 & \CC 21.9569 \\
Concept-Art & $\mathrm{cos}(\frac{\pi t}{2})$ & 0.2928 & 1.0219 & 6.3508 & \CC 21.8251 \\
Concept-Art & $1-t^{0.6}$ & \CC 0.2937 & \CC 1.0238 & \CC 6.3602 & 21.8232 \\
\bottomrule
\end{tabular}}
\caption{\small {Quantitative comparison between cosine schedule and $1-t^{0.6}$ on the HPD v2.}}
\label{tab:inv_karras_comparison_fullHPD}
\vskip -0.01in
\end{table}

%% file: tables/zigzag_sampling_apd_1.tex
\begin{table}[!t]
\centering
\vskip -0.01in
\small
\renewcommand\arraystretch{1.2}
\setlength{\tabcolsep}{10pt}
\hspace{-0.5cm}
\scalebox{0.58}{
\begin{tabular}{cccccc}
\hline
Method & PickScore (↑) & HPS v2 (↑) & AES (↑) & ImageReward (↑) & CLIPScore (↑) \\ \hline
Vanilla Sampling & \CC 22.5034 & \CC 0.30566 & \CC 6.2038 & \CC 1.0523 & 0.8378 \\
Z-Sampling (8) & 22.4932 & 0.30476 & 6.1949 & 1.0344 & \CC 0.8398 \\
Z-Sampling (16) & 22.4295 & 0.27625 & 6.1669 & 1.0133 & 0.8317 \\
Z-Sampling (32) & 22.0241 & 0.29795 & 6.0329 & 0.8596 & 0.8185 \\
Z-Sampling (48) & 21.3509 & 0.30499 & 5.7563 & 0.6134 & 0.8025 \\
Z-Sampling (64) & 18.1041 & 0.11721 & 4.0962 & -2.0106 & 0.6003 \\
\bottomrule
\end{tabular}
}
\caption{\small {The experimental result of Z-Sampling using the random masking mechanism.}}
\label{tab:random_masking_mechanism}
\vskip -0.01in
\end{table}

%% file: tables/zigzag_sampling_apd_2.tex
\begin{table}[ht]
\centering
\resizebox{0.48\textwidth}{!}{%
\begin{tabular}{cccccc}
\hline
{Sampling Step $N$} & {Method} & {PickScore (↑)} & {HPS v2 (↑)} & {AES (↑)} & {ImageReward (↑)} \\
\hline
\multirow{3}{*}{16} & Vanilla Sampling & 21.7214 & 0.2792 & 5.9680 & 0.7154 \\
                     & Recover Z-Sampling & 21.9711 & 0.2853 & 6.0249 & 0.8400 \\
                     & Masked Z-Sampling & \CC 22.2514 & \CC 0.3016 & \CC 6.1017 & \CC 0.8901 \\
\hline
\multirow{3}{*}{48} & Vanilla Sampling & 22.4894 & 0.3048 & 6.2157 & 1.0804 \\
                     & Recover Z-Sampling & 22.4380 & 0.3041 & 6.1774 & 1.0242 \\
                     & Masked Z-Sampling & \CC 22.5826 & \CC 0.3110 & \CC 6.2198 & \CC 1.0918 \\
\hline
\multirow{3}{*}{64} & Vanilla Sampling & 22.5034 & 0.3056 & 6.2038 & 1.0523 \\
                     & Recover Z-Sampling & 22.4747 & 0.3064 & \CC 6.2296 & \CC 1.0838 \\
                     & Masked Z-Sampling & \CC 22.5375 & \CC 0.3087 & 6.1769 & 1.0559 \\
\hline
\end{tabular}}
\caption{Comparison of vanilla sampling, masked Z-Sampling and recover Z-Sampling on the HPD v2 Subset.}
\label{tab:zigzag_sampling_recover_improved}
\end{table}

%% file: tables/noise_regularization_apd_1.tex
\begin{table}[ht]
\centering
\resizebox{0.48\textwidth}{!}{%
\begin{tabular}{lcccccc}
\hline
{Dataset} & {Method} & {PickScore (↑)} & {HPS v2 (↑)} & {AES (↑)} & {ImageReward (↑)} \\
\hline
\multirow{2}{*}{Anime} & Vanilla Sampling & 22.4776 & 0.3053 & 6.1049 & 0.9577 \\
                             & Noise Regularization & \CC 22.6988 & \CC 0.3133 & \CC 6.1262 & \CC 1.1107 \\
\hline
\multirow{2}{*}{Concept-art} & Vanilla Sampling & 21.8251 & 0.2928 & 6.3509 & 1.0219 \\
                                    & Noise Regularization & \CC 21.8862 & \CC 0.2951 & \CC 6.3844 & \CC 1.0492 \\
\hline
\multirow{2}{*}{Photo} & Vanilla Sampling & 21.5552 & 0.2658 & 5.9993 & 0.5161 \\
                              & Noise Regularization & \CC 21.6358 & \CC 0.2691 & \CC 6.0055 & \CC 0.6005 \\
\hline
\multirow{2}{*}{Paintings} & Vanilla Sampling & 21.8669 & 0.2915 & 6.4791 & 1.0802 \\
                                  & Noise Regularization & \CC 21.9340 & \CC 0.2928 & \CC 6.4997 & \CC 1.1013 \\
\hline
\end{tabular}}
\caption{Comparison of vanilla sampling and noise regularization on HPD v2.}
\label{tab:noise_regularization_comparison}
\end{table}

%% file: tables/differential_sampling_apd_1.tex
\begin{table}[ht]
\centering
\resizebox{0.48\textwidth}{!}{%
\begin{tabular}{lcccccc}
\hline
{Dataset} & {Method} & {PickScore (↑)} & {HPSV2 (↑)} & {AES (↑)} & {ImageReward (↑)} \\
\hline
\multirow{2}{*}{Anime} & Vanilla Sampling & 22.4776 & 0.3053 & 6.1049 & 0.9577 \\
                              & Differential Sampling & \CC 22.6472 & \CC 0.3108 & \CC 6.1050 & \CC 1.0796 \\
\hline
\multirow{2}{*}{Photo} & Vanilla Sampling & 21.5552 & 0.2658 & 5.9992 & 0.5161 \\
                              & Differential Sampling & \CC 21.6730 & \CC 0.2700 & \CC 6.0203 & \CC 0.6067 \\
\hline
\multirow{2}{*}{Paintings} & Vanilla Sampling & 21.8669 & 0.2915 &  \CC 6.4790 & 1.0802 \\
                                  & Differential Sampling & \CC 21.9769 & \CC 0.2917 & 6.4592 & \CC 1.0944 \\
\hline
\multirow{2}{*}{Concept-Art} & Vanilla Sampling & 21.8251 & 0.2928 & 6.3508 & 1.0219 \\
                                  & Differential Sampling & \CC 21.8630 & \CC 0.2951 & \CC 6.3659 & \CC 1.0493 \\
\hline
\end{tabular}}
\caption{Comparison of vanilla sampling and differential sampling on HPD v2.}
\label{tab:differential_sampling_comparison}
\end{table}

%% file: tables/tomemgt_apd_1.tex
\begin{table}[ht]
\centering
\resizebox{0.48\textwidth}{!}{%
\begin{tabular}{lccccc}
\hline
{Dataset} & {Method} & {HPS v2 (↑)} & {ImageReward (↑)} & {AES (↑)} & {PickScore (↑)} \\
\hline
\multirow{2}{*}{Anime} & Vanilla Sampling & \CC 0.3053 &\CC 0.9577 &\CC 6.1049 & \CC 22.4776 \\
                              & TomeMGT & 0.2866 & 0.7919 & 5.8713 & 21.8028 \\
\hline
\multirow{2}{*}{Photo} & Vanilla Sampling & \CC 0.2658 & \CC 0.5161 & \CC 5.9992 & \CC 21.5552 \\
                              & TomeMGT & 0.2430 & 0.1883 & 5.7982 & 20.8134 \\
\hline
\multirow{2}{*}{Paintings} & Vanilla Sampling & \CC 0.2915 & \CC 1.0802 & \CC 6.4790 & \CC 21.8669 \\
                                  & TomeMGT & 0.2736 & 0.7780 & 6.1859 & 21.0760 \\
\hline
\multirow{2}{*}{Concept-Art} & Vanilla Sampling & \CC 0.2928 & \CC 1.0219 & \CC 6.3508 & \CC 21.8251 \\
                                  & TomeMGT & 0.2757 & 0.7658 & 6.0891 & 21.0819 \\
\hline
\end{tabular}}
\caption{Comparison between TomeMGT and vanilla sampling.}
\label{tab:tomemgt_comparison}
\end{table}

%% file: tables/meissonic_512_512_combination.tex
\begin{table*}[!b]
\vspace{-2pt}
\centering
\renewcommand\arraystretch{1.0}
\resizebox{0.95\textwidth}{!}{%
\begin{tabular}{lccccccc}
\hline
{Dataset} & {Noise Regularization} & {Differential Sampling} & {Masked Z-Sampling} & {PickScore ($\uparrow$)} & {HPS v2 ($\uparrow$)} & {AES ($\uparrow$)} & {ImageReward ($\uparrow$)} \\
\hline
\multirow{4}{*}{Anime} &  &  &  & 22.5610 & 0.3139 & 5.9191 & 1.1811 \\
 & \cmark &  &  & 22.5734 (53\%) & 0.3152 (51\%) & 5.9066 (47\%) & 1.1779 (50\%) \\
 &  &  & \cmark & 22.6275 (56\%) & 0.3157 (52\%) & 5.9381 (56\%) & 1.1967 (54\%) \\
 & \cmark & \cmark & \cmark & \CC 22.6528 (59\%) & \CC 0.3175 (56\%) & \CC 5.9169 (50\%) &\CC 1.1994 (51\%) \\
\hline
\multirow{4}{*}{Photo} &  &  &  & 21.6864 & 0.2762 & 5.8391 & 0.6886 \\
 & \cmark &  &  & 21.7291 (53\%) & 0.2789 (53\%) & 5.8550 (52\%) & 0.7278 (52\%) \\
 &  & \cmark & \cmark & \CC 21.7558 (54\%) & \CC 0.2815 (59\%) & \CC 5.8782 (56\%) & \CC 0.7974 (58\%) \\
 & \cmark & \cmark & \cmark & 21.7528 (54\%) & 0.2792 (56\%) & 5.8741 (53\%) & 0.7937 (57\%) \\
\hline
\multirow{4}{*}{Concept-art} &  &  &  & 21.7960 & 0.2949 & 6.2479 & 1.1199 \\
 & \cmark &  &  & 21.8513 (52\%) & 0.2979 (56\%) & 6.2383 (47\%) & 1.1274 (50\%) \\
 &  &  & \cmark & 21.8882 (59\%) & 0.2982 (60\%) & \CC 6.2554 (51\%) & 1.1401 (54\%) \\
 & \cmark & \cmark & \cmark & \CC 21.9483 (62\%) &\CC  0.3013 (62\%) & 6.2370 (50\%) & \CC 1.1545 (52\%) \\
\hline
\multirow{4}{*}{Paintings} &  &  &  & 21.8358 & 0.2924 & 6.3968 & 1.1747 \\
 & \cmark &  &  & 21.8839 (52\%) & 0.2941 (55\%) & \CC 6.3992 (49\%) & \CC 1.1855 (50\%) \\
 &  & \cmark & \cmark & 21.9344 (57\%) & 0.2954 (58\%) & 6.3914 (50\%) & 1.1839 (53\%) \\
 & \cmark & \cmark & \cmark & \CC 21.9621 (60\%) & \CC 0.2959 (59\%) & 6.4006 (51\%) & 1.1728 (50\%) \\
\hline
\end{tabular}}
\vspace{-5pt}
\caption{Comparison of combination of different design choices with Meissonic-512$\times$512 on HPD v2.}
\label{tab:meissonic_512_evaluation}
\vspace{-8pt}
\end{table*}

%% file: tables/meissonic_1024_1024_combination.tex
\begin{table*}[!b]
\vspace{-2pt}
\centering
\renewcommand\arraystretch{1.0}
\resizebox{0.95\textwidth}{!}{%
\begin{tabular}{lccccccc}
\hline
{Dataset} & {Noise Regularization} & {Differential Sampling} & {Masked Z-Sampling} & {PickScore ($\uparrow$)} & {HPS v2 ($\uparrow$)} & {AES ($\uparrow$)} & {ImageReward ($\uparrow$)} \\
\hline
\multirow{6}{*}{Anime} & \cmark &  &  & 22.6988 (58\%) & 0.3134 (56\%) & 6.1262 (52\%) & 1.1107 (55\%) \\
 &  &  & \cmark & 22.7559 (61\%) & 0.3184 (68\%) & 6.1165 (51\%) & 1.1175 (57\%) \\
 & \cmark &  & \cmark & \CC 22.8144 (65\%) & 0.3199 (69\%) & 6.1364 (55\%) & \CC 1.1343 (58\%) \\
 &  & \cmark & \cmark & 22.8115 (65\%) & \CC 0.3203 (73\%) & \CC 6.1375 (55\%) & 1.1130 (56\%) \\
 & \cmark & \cmark & \cmark & 22.7559 (60\%) & 0.3184 (68\%) & 6.1165 (51\%) & 1.1175 (57\%) \\
\hline
\multirow{6}{*}{Concept-art} & \cmark &  &  & 21.8862 (55\%) & 0.2951 (54\%) & 6.3844 (57\%) & 1.0492 (52\%) \\
 &  &  & \cmark & 21.9617 (59\%) & 0.2996 (65\%) & 6.3679 (51\%) & 1.0611 (55\%) \\
 & \cmark &  & \cmark & \CC 22.0411 (65\%) &\CC 0.3032 (71\%) & \CC 6.4027 (57\%) &  1.0687 (55\%) \\
 &  & \cmark & \cmark & 22.0112 (64\%) & 0.3024 (70\%) & 6.3857 (55\%) & \CC 1.0720 (67\%) \\
 & \cmark & \cmark & \cmark & 21.9617 (59\%) & 0.2996 (65\%) & 6.3679 (51\%) & 1.0611 (55\%) \\
\hline
\multirow{6}{*}{Photo} & \cmark &  &  & 21.6358 (54\%) & 0.2692 (55\%) & 6.0056 (51\%) & 0.6005 (53\%) \\
 &  &  & \cmark & 21.6628 (55\%) & 0.2747 (65\%) & 6.0728 (60\%) & 0.6307 (58\%) \\
 & \cmark &  & \cmark & \CC 21.7010 (59\%) & \CC 0.2768 (69\%) & \CC 6.0898 (61\%) & \CC 0.6897 (60\%) \\
 &  & \cmark & \cmark & 21.6634 (58\%) & 0.2753 (67\%) &  6.0805 (62\%) & 0.6427 (57\%) \\
 & \cmark & \cmark & \cmark & 21.6628 (55\%) & 0.2747 (65\%) & 6.0728 (60\%) & 0.6307 (58\%) \\
\hline
\multirow{6}{*}{Paintings} & \cmark &  &  & 21.9340 (55\%) & 0.2928 (52\%) & 6.4998 (55\%) & 1.1013 (52\%) \\
 &  &  & \cmark & 22.0255 (62\%) & 0.2977 (62\%) & 6.5036 (54\%) & 1.1193 (53\%) \\
 & \cmark &  & \cmark &  22.0595 (64\%) & \CC 0.2995 (69\%) & \CC 6.5164 (58\%) & \CC 1.1249 (55\%) \\
 &  & \cmark & \cmark & \CC 22.0619 (60\%) & 0.2989 (67\%) & 6.4808 (52\%) & 1.1067 (55\%) \\
 & \cmark & \cmark & \cmark & 22.0255 (62\%) & 0.2977 (62\%) & 6.5036 (54\%) & 1.1193 (53\%) \\
\hline
\end{tabular}}
\vspace{-5pt}
\caption{Comparison of combination of different design choices with Meissonic-1024$\times$1024 on HPD v2.}
\label{tab:meissonic_1024_evaluation}
\vspace{-8pt}
\end{table*}

%% file: tables/challengebench_appendix.tex
\begin{table*}[!b]
\vspace{-8pt}
\centering
\renewcommand\arraystretch{1.0}
\resizebox{0.85\textwidth}{!}{%
\begin{tabular}{llcccc}
\hline
{Model} & {Method} & {HPS v2 (↑)} & {ImageReward (↑)} & {AES (↑)} & {PickScore (↑)} \\
\hline
Meissonic & Vanilla Sampling & 0.2116 & 0.0670 & 5.8503 & 19.3237 \\
Meissonic & Differential Sampling & 0.2135 & 0.1114 & 5.8957 & 19.3381 \\
Meissonic & Our Noise Schedule & 0.2120 & 0.0473 & 5.8635 & 19.3147 \\
Meissonic & Masked Z-Sampling & 0.2183 & 0.1436 & 5.8885 & 19.3926 \\
Meissonic & Noise Regularization & 0.2128 & 0.0748 & 5.8888 & 19.3645 \\
Meissonic & TokenMGT ($z$\%=50\%)& 0.2000 & -0.2593 & 5.5003 & 18.9691 \\
Meissonic & SCQ (W4A8) & 0.2114 & 0.0590 & 5.8431 & 19.3365 \\
SD XL & N/A & 0.1838 & -0.4915 & 5.8195 & 19.5792 \\
\makecell[l]{FLUX.1-schnell (quant 8bit)} & N/A & \CC 0.2257 & \CC 0.2595 & 6.0357 & 19.6465 \\
\makecell[l]{SD-3.5-Large (quant 4bit)} & N/A & 0.2114 & 0.0373 & \CC 6.0848 & \CC 19.7922 \\
\hline
\end{tabular}}
\vspace{-5pt}
\caption{Comparison of various models and methods on Challengebench.}
\label{tab:challengebench_apd}
\vspace{-10pt}
\end{table*}

%% file: tables/llamagen.tex
\begin{table*}[!h]
\vspace{-2pt}
\begin{center}
\footnotesize
\renewcommand\arraystretch{0.85}
\setlength{\tabcolsep}{10pt}
\vspace{-0.1in}
\resizebox{1.0\linewidth}{!}{\begin{tabular}{cccccccc}
\hline
              GenEval         & \multirow{2}{*}{Single Object~(↑)} & \multirow{2}{*}{Two Object~(↑)} & \multirow{2}{*}{Counting~(↑)} & \multirow{2}{*}{Colors~(↑)} & \multirow{2}{*}{Position~(↑)} & \multirow{2}{*}{Color Attr~(↑)} & \multirow{2}{*}{Avg.~(↑)}     \\
(LlamaGen-512$\times$512) &  & & & & & &  \\ \hline
Vanilla Sampling & \CC {20.00\%} & 9.09\% & 1.25\% & \CC {15.96\%} & 11.00\% & 0.00\% & 9.55\% \\
Noise Regularization & \CC {20.00\%} & 9.09\% & \CC {2.50\%} & 13.83\% & \CC {16.00\%} & 1.00\% & 10.40\% \\
Differential Sampling & \CC {20.00\%} & \CC {12.12\%} & \CC {2.50\%} & 14.26\% & 14.00\% &\CC { 2.00\%} & \CC {10.81\%} \\ \hline
\end{tabular}
}
\end{center}
\vspace{-8pt}
\caption{Comparison of our proposed methods on GenEval~\citep{geneval} with LlamaGen-512$\times$512.}
\label{tab:llamagen_1}
\vspace{-7pt}
\end{table*}\begin{table*}[!h]
\vspace{-2pt}
\begin{center}
\footnotesize
\renewcommand\arraystretch{0.65}
\setlength{\tabcolsep}{10pt}
\vspace{-0.1in}
\resizebox{0.75\linewidth}{!}{
{\begin{tabular}{cccccc}
\hline
LlamaGen-256$\times$256 & IS~(↑) & FID~(↓) & Prec.~(↑) & Recall~(↑) & sFID~(↓) \\ 
\hline
Vanilla Sampling & 316.68 & 13.22 & 0.8723 & \CC {0.215} & 18.10 \\
Noise Regularization & 317.41 & 13.02 & \CC {0.8741} & 0.207 & 17.44 \\
Differential Sampling & \CC {317.80 }& \CC {12.87} & 0.8732 & 0.206 & \CC {15.87} \\
\hline
\end{tabular}}
}
\end{center}
\vspace{-8pt}
\caption{Comparison of our proposed methods on the traditional metrics with LlamaGen-256$\times$256.}
\label{tab:llamagen_2}
\vspace{-7pt}
\end{table*}

%% file: main.bbl
\begin{thebibliography}{55}
\providecommand{\natexlab}[1]{#1}
\providecommand{\url}[1]{\texttt{#1}}
\expandafter\ifx\csname urlstyle\endcsname\relax
  \providecommand{\doi}[1]{doi: #1}\else
  \providecommand{\doi}{doi: \begingroup \urlstyle{rm}\Url}\fi

\bibitem[Achiam et~al.(2023)Achiam, Adler, Agarwal, Ahmad, Akkaya, Aleman, Almeida, Altenschmidt, Altman, Anadkat, et~al.]{GPT4}
Josh Achiam, Steven Adler, Sandhini Agarwal, Lama Ahmad, Ilge Akkaya, Florencia~Leoni Aleman, Diogo Almeida, Janko Altenschmidt, Sam Altman, Shyamal Anadkat, et~al.
\newblock Gpt-4 technical report.
\newblock \emph{arXiv preprint arXiv:2303.08774}, 2023.

\bibitem[Anonymous(2024)]{zigzag}
Anonymous.
\newblock Zigzag diffusion sampling: The path to success ls zigzag.
\newblock In \emph{Submitted to The Thirteenth International Conference on Learning Representations}, 2024.
\newblock under review.

\bibitem[Bai et~al.(2024)Bai, Ye, Chow, Song, Chen, Li, Dong, Zhu, and Yan]{meissonic}
Jinbin Bai, Tian Ye, Wei Chow, Enxin Song, Qing-Guo Chen, Xiangtai Li, Zhen Dong, Lei Zhu, and Shuicheng Yan.
\newblock Meissonic: Revitalizing masked generative transformers for efficient high-resolution text-to-image synthesis.
\newblock \emph{arXiv preprint arXiv:2410.08261}, 2024.

\bibitem[Bolya and Hoffman(2023)]{tomesd}
Daniel Bolya and Judy Hoffman.
\newblock Token merging for fast stable diffusion.
\newblock In \emph{Proceedings of the IEEE/CVF conference on computer vision and pattern recognition}, pages 4599--4603, 2023.

\bibitem[Chang et~al.(2022)Chang, Zhang, Jiang, Liu, and Freeman]{maskgit}
Huiwen Chang, Han Zhang, Lu Jiang, Ce Liu, and William~T Freeman.
\newblock Maskgit: Masked generative image transformer.
\newblock In \emph{Proceedings of the IEEE/CVF Conference on Computer Vision and Pattern Recognition}, pages 11315--11325, 2022.

\bibitem[Chang et~al.(2023)Chang, Zhang, Barber, Maschinot, Lezama, Jiang, Yang, Murphy, Freeman, Rubinstein, et~al.]{muse}
Huiwen Chang, Han Zhang, Jarred Barber, AJ Maschinot, Jose Lezama, Lu Jiang, Ming-Hsuan Yang, Kevin Murphy, William~T Freeman, Michael Rubinstein, et~al.
\newblock Muse: Text-to-image generation via masked generative transformers.
\newblock \emph{arXiv preprint arXiv:2301.00704}, 2023.

\bibitem[Chen et~al.(2024)Chen, Meng, Tang, Ma, Jiang, Wang, Wang, and Zhu]{QDIT}
Lei Chen, Yuan Meng, Chen Tang, Xinzhu Ma, Jingyan Jiang, Xin Wang, Zhi Wang, and Wenwu Zhu.
\newblock Q-dit: Accurate post-training quantization for diffusion transformers.
\newblock \emph{arXiv preprint arXiv:2406.17343}, 2024.

\bibitem[Fan et~al.(2024)Fan, Li, Qin, Li, Sun, Rubinstein, Sun, He, and Tian]{fluid}
Lijie Fan, Tianhong Li, Siyang Qin, Yuanzhen Li, Chen Sun, Michael Rubinstein, Deqing Sun, Kaiming He, and Yonglong Tian.
\newblock Fluid: Scaling autoregressive text-to-image generative models with continuous tokens.
\newblock \emph{arXiv preprint arXiv:2410.13863}, 2024.

\bibitem[Ghosh et~al.(2023)Ghosh, Hajishirzi, and Schmidt]{geneval}
Dhruba Ghosh, Hannaneh Hajishirzi, and Ludwig Schmidt.
\newblock Geneval: An object-focused framework for evaluating text-to-image alignment.
\newblock \emph{Advances in Neural Information Processing Systems}, 36:\penalty0 52132--52152, 2023.

\bibitem[Hessel et~al.(2022)Hessel, Holtzman, Forbes, Bras, and Choi]{CLIPScore}
Jack Hessel, Ari Holtzman, Maxwell Forbes, Ronan~Le Bras, and Yejin Choi.
\newblock Clipscore: A reference-free evaluation metric for image captioning, 2022.

\bibitem[Heusel et~al.(2017)Heusel, Ramsauer, Unterthiner, Nessler, and Hochreiter]{fid}
Martin Heusel, Hubert Ramsauer, Thomas Unterthiner, Bernhard Nessler, and Sepp Hochreiter.
\newblock Gans trained by a two time-scale update rule converge to a local nash equilibrium.
\newblock In \emph{Neural Information Processing Systems}, Long Beach Convention Center, Long Beach, 2017. NeurIPS.

\bibitem[Ho and Salimans(2021)]{nips2021_classifier_free_guidance}
Jonathan Ho and Tim Salimans.
\newblock Classifier-free diffusion guidance.
\newblock In \emph{Neural Information Processing Systems Workshop}, Virtual Event, 2021. NeurIPS.

\bibitem[Ho et~al.(2020)Ho, Jain, and Abbeel]{ddpm_begin}
Jonathan Ho, Ajay Jain, and Pieter Abbeel.
\newblock Denoising diffusion probabilistic models.
\newblock In \emph{Neural Information Processing Systems}, pages 6840--6851, Virtual Event, 2020. NeurIPS.

\bibitem[Huang et~al.(2023)Huang, Sun, Xie, Li, and Liu]{t2i_compbench}
Kaiyi Huang, Kaiyue Sun, Enze Xie, Zhenguo Li, and Xihui Liu.
\newblock T2i-compbench: A comprehensive benchmark for open-world compositional text-to-image generation.
\newblock \emph{Advances in Neural Information Processing Systems}, 36:\penalty0 78723--78747, 2023.

\bibitem[Jacob et~al.(2018)Jacob, Kligys, Chen, Zhu, Tang, Howard, Adam, and Kalenichenko]{jacob2018quantization}
Benoit Jacob, Skirmantas Kligys, Bo Chen, Menglong Zhu, Matthew Tang, Andrew Howard, Hartwig Adam, and Dmitry Kalenichenko.
\newblock Quantization and training of neural networks for efficient integer-arithmetic-only inference.
\newblock In \emph{Proceedings of the IEEE conference on computer vision and pattern recognition}, pages 2704--2713, 2018.

\bibitem[Karras et~al.(2022)Karras, Aittala, Aila, and Laine]{karras2022elucidating}
Tero Karras, Miika Aittala, Timo Aila, and Samuli Laine.
\newblock Elucidating the design space of diffusion-based generative models.
\newblock \emph{Advances in neural information processing systems}, 35:\penalty0 26565--26577, 2022.

\bibitem[Kingma(2013)]{VAE}
Diederik~P Kingma.
\newblock Auto-encoding variational bayes.
\newblock \emph{arXiv preprint arXiv:1312.6114}, 2013.

\bibitem[Labs(2024)]{FLUX}
Black~Forest Labs.
\newblock Flux.
\newblock \url{https://blackforestlabs.ai/}, 2024.

\bibitem[Laion.ai(2022)]{AES}
Laion.ai.
\newblock Laion-aesthetics.
\newblock \url{https://laion.ai/blog/laion-aesthetics/}, 2022.

\bibitem[Lee et~al.(2023)Lee, Kim, and Ye]{icml23_curvature}
Sangyun Lee, Beomsu Kim, and Jong~Chul Ye.
\newblock Minimizing trajectory curvature of ode-based generative models.
\newblock \emph{arXiv preprint arXiv:2301.12003}, 2023.

\bibitem[Lin et~al.(2014)Lin, Maire, Belongie, Hays, Perona, Ramanan, Doll{\'a}r, and Zitnick]{COCO}
Tsung-Yi Lin, Michael Maire, Serge Belongie, James Hays, Pietro Perona, Deva Ramanan, Piotr Doll{\'a}r, and C~Lawrence Zitnick.
\newblock Microsoft coco: Common objects in context.
\newblock In \emph{European Conference on Computer Vision}, pages 740--755. Springer, 2014.

\bibitem[Liu et~al.(2024{\natexlab{a}})Liu, Shao, Li, Bai, Xiong, Kwok, Helal, and Xie]{liu2024alignment}
Buhua Liu, Shitong Shao, Bao Li, Lichen Bai, Haoyi Xiong, James Kwok, Sumi Helal, and Zeke Xie.
\newblock Alignment of diffusion models: Fundamentals, challenges, and future.
\newblock \emph{arXiv preprint arXiv:2409.07253}, 2024{\natexlab{a}}.

\bibitem[Liu et~al.(2024{\natexlab{b}})Liu, Zhao, Zhuo, Lin, Qiao, Li, and Gao]{luminia_mgpt}
Dongyang Liu, Shitian Zhao, Le Zhuo, Weifeng Lin, Yu Qiao, Hongsheng Li, and Peng Gao.
\newblock Lumina-mgpt: Illuminate flexible photorealistic text-to-image generation with multimodal generative pretraining.
\newblock \emph{arXiv preprint arXiv:2408.02657}, 2024{\natexlab{b}}.

\bibitem[Liu et~al.(2022)Liu, Gong, and Liu]{iclr22_rect}
Xingchao Liu, Chengyue Gong, and Qiang Liu.
\newblock Flow straight and fast: Learning to generate and transfer data with rectified flow.
\newblock \emph{arXiv preprint arXiv:2209.03003}, 2022.

\bibitem[Lu et~al.(2022{\natexlab{a}})Lu, Zhou, Bao, Chen, and Li]{dpm_solver++}
Cheng Lu, Yuhao Zhou, Fan Bao, Jianfei Chen, and Chongxuan Li.
\newblock Dpm-solver++: Fast solver for guided sampling of diffusion probabilistic models.
\newblock \emph{arXiv preprint arXiv:2211.01095}, 2022{\natexlab{a}}.

\bibitem[Lu et~al.(2022{\natexlab{b}})Lu, Zhou, Bao, Chen, Li, and Zhu]{dpm_solver}
Cheng Lu, Yuhao Zhou, Fan Bao, Jianfei Chen, Chongxuan Li, and Jun Zhu.
\newblock Dpm-solver: A fast ode solver for diffusion probabilistic model sampling in around 10 steps.
\newblock In \emph{Neural Information Processing Systems}, New Orleans, LA, USA, 2022{\natexlab{b}}. NeurIPS.

\bibitem[Meng et~al.(2022)Meng, Gao, Kingma, Ermon, Ho, and Salimans]{cvpr22_kd_guided}
Chenlin Meng, Ruiqi Gao, Diederik~P Kingma, Stefano Ermon, Jonathan Ho, and Tim Salimans.
\newblock On distillation of guided diffusion models.
\newblock \emph{arXiv preprint arXiv:2210.03142}, 2022.

\bibitem[Mokady et~al.(2023)Mokady, Hertz, Aberman, Pritch, and Cohen-Or]{ddiminversion}
Ron Mokady, Amir Hertz, Kfir Aberman, Yael Pritch, and Daniel Cohen-Or.
\newblock Null-text inversion for editing real images using guided diffusion models.
\newblock In \emph{Proceedings of the IEEE/CVF Conference on Computer Vision and Pattern Recognition}, pages 6038--6047, 2023.

\bibitem[OpenAI(2024)]{strawberry}
OpenAI.
\newblock Learning to reason with llms, 2024.

\bibitem[Peebles and Xie(2023)]{DIT}
William Peebles and Saining Xie.
\newblock Scalable diffusion models with transformers.
\newblock In \emph{Proceedings of the IEEE/CVF International Conference on Computer Vision}, pages 4195--4205, 2023.

\bibitem[Podell et~al.()Podell, English, Lacey, Blattmann, Dockhorn, M{\"u}ller, Penna, and Rombach]{SDXL}
Dustin Podell, Zion English, Kyle Lacey, Andreas Blattmann, Tim Dockhorn, Jonas M{\"u}ller, Joe Penna, and Robin Rombach.
\newblock Sdxl: Improving latent diffusion models for high-resolution image synthesis.
\newblock In \emph{The Twelfth International Conference on Learning Representations}.

\bibitem[Qi et~al.(2024)Qi, Bai, Xiong, et~al.]{qi2024not}
Zipeng Qi, Lichen Bai, Haoyi Xiong, et~al.
\newblock Not all noises are created equally: Diffusion noise selection and optimization.
\newblock \emph{arXiv preprint arXiv:2407.14041}, 2024.

\bibitem[Radford et~al.(2021)Radford, Kim, Hallacy, Ramesh, Goh, Agarwal, Sastry, Askell, Mishkin, Clark, Krueger, and Sutskever]{CLIP}
Alec Radford, Jong~Wook Kim, Chris Hallacy, Aditya Ramesh, Gabriel Goh, Sandhini Agarwal, Girish Sastry, Amanda Askell, Pamela Mishkin, Jack Clark, Gretchen Krueger, and Ilya Sutskever.
\newblock Learning transferable visual models from natural language supervision, 2021.

\bibitem[Salimans and Ho(2022)]{iclr22_progressive}
Tim Salimans and Jonathan Ho.
\newblock Progressive distillation for fast sampling of diffusion models.
\newblock In \emph{International Conference on Learning Representations}, Virtual Event, 2022. OpenReview.net.

\bibitem[Salimans et~al.(2016)Salimans, Goodfellow, Zaremba, Cheung, Radford, and Chen]{is}
Tim Salimans, Ian Goodfellow, Wojciech Zaremba, Vicki Cheung, Alec Radford, and Xi Chen.
\newblock Improved techniques for training gans.
\newblock In \emph{Neural Information Processing Systems}, Centre Convencions Internacional Barcelona, Barcelona SPAIN, 2016. NeurIPS.

\bibitem[Sennrich(2015)]{sennrich2015neural}
Rico Sennrich.
\newblock Neural machine translation of rare words with subword units.
\newblock \emph{arXiv preprint arXiv:1508.07909}, 2015.

\bibitem[Shao et~al.(2023{\natexlab{a}})Shao, Dai, Yin, Li, Chen, and Hu]{shao2023catch}
Shitong Shao, Xu Dai, Shouyi Yin, Lujun Li, Huanran Chen, and Yang Hu.
\newblock Catch-up distillation: You only need to train once for accelerating sampling.
\newblock \emph{arXiv preprint arXiv:2305.10769}, 2023{\natexlab{a}}.

\bibitem[Shao et~al.(2023{\natexlab{b}})Shao, Yuan, Huang, Qiu, Wang, and Zhou]{shao2023diffuseexpand}
Shitong Shao, Xiaohan Yuan, Zhen Huang, Ziming Qiu, Shuai Wang, and Kevin Zhou.
\newblock Diffuseexpand: Expanding dataset for 2d medical image segmentation using diffusion models.
\newblock \emph{arXiv preprint arXiv:2304.13416}, 2023{\natexlab{b}}.

\bibitem[Song et~al.(2023{\natexlab{a}})Song, Meng, and Ermon]{ddim}
Jiaming Song, Chenlin Meng, and Stefano Ermon.
\newblock Denoising diffusion implicit models.
\newblock In \emph{International Conference on Learning Representations}, kigali, rwanda, 2023{\natexlab{a}}. OpenReview.net.

\bibitem[Song et~al.(2023{\natexlab{b}})Song, Dhariwal, Chen, and Sutskever]{icml23_consistency}
Yang Song, Prafulla Dhariwal, Mark Chen, and Ilya Sutskever.
\newblock Consistency models.
\newblock \emph{arXiv preprint arXiv:2303.01469}, 2023{\natexlab{b}}.

\bibitem[Song et~al.(2023{\natexlab{c}})Song, Sohl-Dickstein, Kingma, Kumar, Ermon, and Poole]{sde}
Yang Song, Jascha Sohl-Dickstein, Diederik~P Kingma, Abhishek Kumar, Stefano Ermon, and Ben Poole.
\newblock Score-based generative modeling through stochastic differential equations.
\newblock In \emph{International Conference on Learning Representations}, kigali, rwanda, 2023{\natexlab{c}}. OpenReview.net.

\bibitem[Stability.ai(2024)]{SD35}
Stability.ai.
\newblock Introducing stable diffusion 3.5.
\newblock \url{https://stability.ai/news/introducing-stable-diffusion-3-5}, 2024.

\bibitem[Su et~al.(2024)Su, Ahmed, Lu, Pan, Bo, and Liu]{RoPE}
Jianlin Su, Murtadha Ahmed, Yu Lu, Shengfeng Pan, Wen Bo, and Yunfeng Liu.
\newblock Roformer: Enhanced transformer with rotary position embedding.
\newblock \emph{Neurocomputing}, 568:\penalty0 127063, 2024.

\bibitem[Sun et~al.(2024)Sun, Jiang, Chen, Zhang, Peng, Luo, and Yuan]{llamagen}
Peize Sun, Yi Jiang, Shoufa Chen, Shilong Zhang, Bingyue Peng, Ping Luo, and Zehuan Yuan.
\newblock Autoregressive model beats diffusion: Llama for scalable image generation.
\newblock \emph{arXiv preprint arXiv:2406.06525}, 2024.

\bibitem[Team et~al.(2024)Team, Mesnard, Hardin, Dadashi, Bhupatiraju, Pathak, Sifre, Rivi{\`e}re, Kale, Love, et~al.]{gemma}
Gemma Team, Thomas Mesnard, Cassidy Hardin, Robert Dadashi, Surya Bhupatiraju, Shreya Pathak, Laurent Sifre, Morgane Rivi{\`e}re, Mihir~Sanjay Kale, Juliette Love, et~al.
\newblock Gemma: Open models based on gemini research and technology.
\newblock \emph{arXiv preprint arXiv:2403.08295}, 2024.

\bibitem[Teng et~al.(2024{\natexlab{a}})Teng, Shi, Liu, Ning, Dai, Wang, Li, and Liu]{jacobidecoding}
Yao Teng, Han Shi, Xian Liu, Xuefei Ning, Guohao Dai, Yu Wang, Zhenguo Li, and Xihui Liu.
\newblock Accelerating auto-regressive text-to-image generation with training-free speculative jacobi decoding.
\newblock \emph{arXiv preprint arXiv:2410.01699}, 2024{\natexlab{a}}.

\bibitem[Teng et~al.(2024{\natexlab{b}})Teng, Shi, Liu, Ning, Dai, Wang, Li, and Liu]{teng2024accelerating}
Yao Teng, Han Shi, Xian Liu, Xuefei Ning, Guohao Dai, Yu Wang, Zhenguo Li, and Xihui Liu.
\newblock Accelerating auto-regressive text-to-image generation with training-free speculative jacobi decoding.
\newblock \emph{arXiv preprint arXiv:2410.01699}, 2024{\natexlab{b}}.

\bibitem[Touvron et~al.(2023)Touvron, Lavril, Izacard, Martinet, Lachaux, Lacroix, Rozi{\`e}re, Goyal, Hambro, Azhar, et~al.]{llama}
Hugo Touvron, Thibaut Lavril, Gautier Izacard, Xavier Martinet, Marie-Anne Lachaux, Timoth{\'e}e Lacroix, Baptiste Rozi{\`e}re, Naman Goyal, Eric Hambro, Faisal Azhar, et~al.
\newblock Llama: Open and efficient foundation language models.
\newblock \emph{arXiv preprint arXiv:2302.13971}, 2023.

\bibitem[Van Den~Oord et~al.(2017)Van Den~Oord, Vinyals, et~al.]{VQVAE}
Aaron Van Den~Oord, Oriol Vinyals, et~al.
\newblock Neural discrete representation learning.
\newblock \emph{Advances in neural information processing systems}, 30, 2017.

\bibitem[Wu et~al.(2023{\natexlab{a}})Wu, Hao, Sun, Chen, Zhu, Zhao, and Li]{HPSV2}
Xiaoshi Wu, Yiming Hao, Keqiang Sun, Yixiong Chen, Feng Zhu, Rui Zhao, and Hongsheng Li.
\newblock Human preference score v2: A solid benchmark for evaluating human preferences of text-to-image synthesis, 2023{\natexlab{a}}.

\bibitem[Wu et~al.(2023{\natexlab{b}})Wu, Hao, Sun, Chen, Zhu, Zhao, and Li]{wu2023human}
Xiaoshi Wu, Yiming Hao, Keqiang Sun, Yixiong Chen, Feng Zhu, Rui Zhao, and Hongsheng Li.
\newblock Human preference score v2: A solid benchmark for evaluating human preferences of text-to-image synthesis.
\newblock \emph{arXiv preprint arXiv:2306.09341}, 2023{\natexlab{b}}.

\bibitem[Xu et~al.(2023)Xu, Liu, Wu, Tong, Li, Ding, Tang, and Dong]{Imagereward}
Jiazheng Xu, Xiao Liu, Yuchen Wu, Yuxuan Tong, Qinkai Li, Ming Ding, Jie Tang, and Yuxiao Dong.
\newblock Imagereward: Learning and evaluating human preferences for text-to-image generation, 2023.

\bibitem[Zhang and Chen(2023)]{zhang2023fast}
Qinsheng Zhang and Yongxin Chen.
\newblock Fast sampling of diffusion models with exponential integrator.
\newblock In \emph{International Conference on Learning Representations}. OpenReview.net, 2023.

\bibitem[Zhao et~al.(2020)Zhao, Jia, and Koltun]{Transformer}
Hengshuang Zhao, Jiaya Jia, and Vladlen Koltun.
\newblock Exploring self-attention for image recognition.
\newblock In \emph{Computer Vision and Pattern Recognition}, pages 10076--10085, 2020.

\bibitem[Zheng et~al.(2023)Zheng, Lu, Chen, and Zhu]{dpm_solver_v3}
Kaiwen Zheng, Cheng Lu, Jianfei Chen, and Jun Zhu.
\newblock Dpm-solver-v3: Improved diffusion ode solver with empirical model statistics.
\newblock \emph{Advances in Neural Information Processing Systems}, 36:\penalty0 55502--55542, 2023.

\end{thebibliography}
